\documentclass[letterpaper, 10 pt, conference]{ieeeconf}  

\IEEEoverridecommandlockouts                              

\usepackage{cite}
\usepackage{amsmath,amssymb,amsfonts,amsthm}
\usepackage[ruled,linesnumbered]{algorithm2e}
\usepackage{algpseudocode}
\usepackage{graphicx}
\usepackage{textcomp}
\usepackage{xcolor}
\usepackage{url}
\usepackage[T1]{fontenc}
\usepackage{cleveref}
\newcounter{subeqn} \renewcommand{\thesubeqn}{\theequation\alph{subeqn}}%
\newcommand{\subeqn}{%
	\refstepcounter{subeqn}
	\tag{\thesubeqn}
}
\newcommand{\R}{\mathbb{R}}

\newcommand{\N}{\mathbb{N}}

\newcommand{\Exp}{\mathbb{E}}
\newcommand{\bx}{\mathbf{x}}
\newcommand{\bu}{\mathbf{u}}

\def\BibTeX{{\rm B\kern-.05em{\sc i\kern-.025em b}\kern-.08em
    T\kern-.1667em\lower.7ex\hbox{E}\kern-.125emX}}
\begin{document}

\title{Non-Gaussian Uncertainty Minimization Based Control of Stochastic Nonlinear Robotic Systems}

\newcommand{\CH}[1]{\textcolor{red}{[CH: #1]}}
\newcommand{\AW}[1]{\textcolor{red}{[AW: #1]}}
\newcommand{\AJ}[1]{\textcolor{red}{[AJ: #1]}}
\let\oldemptyset\emptyset
\let\emptyset\varnothing
\author{Weiqiao Han*, Ashkan Jasour*, Brian Williams
\thanks{Computer Science and Artificial Intelligence Laboratory, Massachusetts Institute of Technology, \{weiqiaoh, jasour, williams\}@mit.edu.
*These authors contributed equally to the paper.}%
}

\maketitle

\begin{abstract}
In this paper, we consider the closed-loop control problem of nonlinear robotic systems in the presence of probabilistic uncertainties and disturbances. More precisely, we design a state feedback controller that minimizes deviations of the states of the system from the nominal state trajectories due to uncertainties and disturbances. Existing approaches to address the control problem of probabilistic systems are limited to particular classes of uncertainties and systems such as Gaussian uncertainties and processes and linearized systems. We present an approach that deals with nonlinear dynamics models and arbitrary known probabilistic uncertainties. We formulate the controller design problem as an optimization problem in terms of statistics of the probability distributions including moments and characteristic functions. In particular, in the provided optimization problem, we use moments and characteristic functions to propagate uncertainties throughout the nonlinear motion model of robotic systems. In order to reduce the tracking deviations, we minimize the uncertainty of the probabilistic states around the nominal trajectory by minimizing the trace and the determinant of the covariance matrix of the probabilistic states. To obtain the state feedback gains, we solve deterministic optimization problems in terms of moments, characteristic functions, and state feedback gains using off-the-shelf interior-point optimization solvers. To illustrate the performance of the proposed method, we compare our method with existing probabilistic control methods.


\end{abstract}
\section{Introduction}
In the control of robotic systems, it is common to first design an open-loop nominal trajectory and then synthesize feedback controllers to follow the nominal trajectory. This standard approach has been applied to vehicle path-following \cite{aguiar2007trajectory}, humanoid robot walking \cite{kuindersma2016optimization}, robotic manipulation \cite{han2020local}, and so on. Usually the  feedback controllers are based on the local linearization around the nominal trajectory. This works well when the state estimation is quite accurate, but might break if there is large uncertainty or external disturbance, and under such unfavorable circumstances, it is important to take uncertainty into consideration and work with stochastic systems. 
In this paper, we consider uncertainty control problems for nonlinear robotic systems subject to probabilistic uncertainties. 
Uncertainty control have many applications in control of robotic and autonomous systems in uncertain environments. This problem is hard and challenging, because it requires
computation and control of time evolution of state uncertainties of the system.

There are several approaches to address control problems of probabilistic systems. These approaches consider probability distributions of the uncertainties to satisfy the control objectives. 
Most existing approaches deal with particular classes of dynamical systems and particular classes of probabilistic uncertainties, notably linear systems and Gaussian uncertainties.
For example, \cite{blackmore2009convex, blackmore2010probabilistic} provide convex linear programs for expected value steering of the states of linear systems in the presence of additive Gaussian uncertainties. Expected value steering based methods do not  control  the  deviation  of  the  states and state uncertainties increase by time. \cite{okamoto2019optimal2,okamoto2019optimal} use convex programs for covariance steering problem of linear systems with additive Gaussian noises. Covariance steering-based methods control the state uncertainties by controlling the covariance of the states. 
The optimal linear covariance steering problem with expected quadratic cost has been solved by \cite{chen2015optimal,chen2015optimal2,chen2018optimal}. 

Most nonlinear covariance steering methods are based on the linearization of the stochastic nonlinear systems. \cite{yi2020nonlinear} developed an efficient algorithm for nonlinear covariance control based on differential dynamic programming (DDP), which relies on local linear models. \cite{ridderhof2019nonlinear} developed an algorithm to solve the nonlinear covariance steering problem by iteratively solving the linear covariance steering problem with respect to the reference trajectory of the previous step, also in the spirit of DDP. \cite{tsolovikos2020nonlinear} deals with systems with unknown dynamics by first fitting a dynamics model and then solves the linear covariance steering control problem at local linear models.

In \cite{jasour2019sequential} chance optimization-based method to design nonlinear state feedback controls for stochastic polynomial systems is provided. The provided approach solves a sequence of semidefinite programs to maximize the probability of remaining in the neighborhood of the nominal trajectory.



\indent\textit{Statement of Contributions}:
In this paper, we present an approach to design state feedback controllers for stochastic nonlinear robotic systems under arbitrary known probabilistic noise to track the nominal trajectory while minimizing uncertainty.
Different from the literature, we directly work with the original nonlinear system, without any linearization. Our approach first computes the statistics (moments) of the system states using exact moment propagation without any approximations, and then searches for state feedback controllers that minimizes the uncertainty at each time step. The feedback gain is obtained by solving nonlinear optimization programs. 
In the experiment section, we compare our method with other probabilistic control methods that could track the nominal trajectory, and we show that our approach outperforms other methods. 
Apart from better performance, our approach has the following advantages. 
First, the approach applies to a general class of stochastic nonlinear systems, more precisely, systems with mixed-trigonometric-polynomial dynamics, which covers most robotic systems, and the approach applies to arbitrary probabilistic uncertainties, not necessarily Gaussian distributions. 
Second, our approach leaves room for nonlinear state feedback control, not necessarily linear state feedback control. 
A disadvantage of our approach is that it requires designing controllers per time step as in \cite{jasour2019sequential}. 
This is the trade-off between performance and engineering effort. 
Note that we assume the open-loop nominal trajectory is given and we design feedback controllers to track the nominal trajectory. 
There are many trajectory optimization methods that could compute the nominal trajectory \cite{han2022non}, and how the nominal trajectory is obtained is not the focus of this paper. 
This assumption is different from most nonlinear covariance steering literature \cite{yi2020nonlinear,ridderhof2019nonlinear,tsolovikos2020nonlinear}, especially methods similar to DDP, where they design a nominal trajectory and feedback controllers together.
We believe that separately designing the nominal trajectory and feedback controllers can be beneficial in many cases.
For instance, the trajectory optimization alone could incorporate more constraints, such as obstacle avoidance, chance constraints, tube safety, and so on, while it is not easy to incorporate more constraints into an algorithm that simultaneously generate nominal trajectories and feedback controllers, especially when dealing with stochastic nonlinear systems. 


\section{Notation and Definitions} \label{sec_def}
This section covers notation and some basic definitions of polynomials and moments of probability distributions.\\

\textbf{Standard Polynomials:} Let {\small $\mathbb{R}[x]$} be the set of real polynomials in the variables {\small $\mathbf{x} \in \mathbb{R}^n$}. Given {\small $ P\in\mathbb{R}[x]$}, we represent {\small $P$} as {\small $\sum_{\alpha\in\mathbb{N}^n} p_\alpha x^\alpha$} using the standard monomial basis $x^\alpha = \Pi_{i=1}^n x_i^{\alpha_i}$ where $\alpha=(\alpha_1,...,\alpha_n) \in \mathbb{N}^n$, and {\small $\mathbf{p}=\{p_\alpha\}_{\alpha\in\mathbb{N}^n}$} denotes the coefficients.

\textbf{Trigonometric Polynomials:} Trigonometric polynomials of order $n$ are defined as $P(x)=\sum_{k=0}^n p_{c_k} \cos(kx)+p_{s_k}\sin(kx) $ where $\{p_{c_k}, p_{s_k}\}_{k=0}^{n}$ are the coefficients\cite{Tri_1,jasour2021moment}.

\textbf{Mixed Trigonometric Polynomials:} Mixed trigonometric polynomials are a mixture of standard and trigonometric polynomials and are defined as  $P(x)=\sum_{k=0}^n p_k x^{a_k}\cos^{b_k}(x)\sin^{c_k}(x)$ where $a_k, b_k, c_k \in \mathbb{N}$ and $\{p_k\}_{k=0}^{n}$ are the coefficients \cite{Mixed_Tri_1, Mixed_Tri_2,jasour2021moment}.\\

In this paper, we will use standard, trigonometric, and mixed trigonometric polynomials to represent nonlinear robotic systems, e.g.,  $x(k+1)=x(k)+v(k)\cos(\theta(k))$, $y(k+1)=y(k)+v(k)\sin(\theta(k))$ where $(x,y)$, $v$, and $\theta$ are position, velocity, and steering angle of a mobile robot, respectively.
\\

\textbf{Moments of Probability Distributions:} Let $(\Omega, \Sigma, \mu)$ be a probability space, where $\Omega$ is the sample space, $\Sigma$ is the $\sigma$-algebra of $\Omega$, and $\mu: \Sigma \to [0,1]$ is the probability measure on $\Sigma$. 
Suppose $\mathbf{x} \in \Omega \subseteq \R^n$ is an $n$-dimensional random vector. 
Let $\mathbf{\alpha} = (\alpha_1,\ldots,\alpha_n)\in \N^n$.
The expectation of $\mathbf{x}^{\mathbf{\alpha}}$ defined as $ \mathbb{E}[\mathbf{x}^{\mathbf{\alpha}}] $ is a {moment of order $\alpha$}, where $\alpha = \sum_i \alpha_i$.
The sequence of all moments of order $\alpha$ is the expectation of all monomials of order $\alpha$ sorted in graded reverse lexicographic order (grevlex).
For example, the sequence of moments of order $\alpha = 2$ of random vector $\mathbf{x} \in \mathbb{R}^2$ is \begin{small}
$[\mathbb{E}[x_1^2], \mathbb{E}[x_1x_2], \mathbb{E}[x_2^2]]$. \end{small} 
One can use first and second order moments to describe the covariance of random variables, i.e.,  $\mathbb{E}[\mathbf{x}^2]-\mathbb{E}^2[\mathbf{x}]$. In this paper, we will use moments of probability distributions to represent uncertainties.

\section{Problem Formulation}

Consider the following stochastic discrete-time continuous-state nonlinear system as

\begin{equation} \label{sys}
\mathbf{x}(k+1)=\mathbf{f}(\mathbf{x}(k),\mathbf{u}(k),\mathbf{\omega}(k))
\end{equation}\noindent where $\mathbf{x}(k) \in \mathbb{R}^{n_x}$ is the state vector, $\mathbf{u}(k) \in \mathbb{R}^{n_u}$ is the input vector, and $\mathbf{\omega}(k) \in \mathbb{R}^{n_{\omega}}$ is the vector of uncertain  parameters,  such  as  uncertain model parameters and external disturbances, with known  probability  distribution $p_{\omega_k}(\omega)$. 

We  assume  that function $\mathbf{f}$ only  contains  certain  elementary  functions,  including  polynomials, trigonometric  functions,  and  mixed trigonometric polynomial  functions. This  assumption is not conservative, given  that  many  robotic  dynamical  systems  can  be  represented  by  these  elementary  functions. 
We also assume that the moments of the uncertainty probability  distribution $p_{\omega_k}(\omega)$ can be calculated analytically or approximated up to certain accuracy, and are finite. For example, we cannot deal with Cauchy distribution, because its moments are infinite.

Let {\small$\mathbf{x}^*:=\{\mathbf{x}^*(k), \ k=1,...,T\}$} be the given nominal trajectory and  {\small$\mathbf{u}^*:=\{\mathbf{u}^*(k), \ k=0,...,T-1\}$} be the given nominal open-loop control input. In the absence of uncertainties, system \eqref{sys} follows the given nominal trajectory {\small$\mathbf{x}^*$} under the nominal open-loop control {\small$\mathbf{u}^*$}. In the presence of uncertainties, states of the system
tend to deviate from the planned trajectory. Hence, to achieve the planning goals, we will design time-varying feedback controllers to steer the deviated states in the neighborhood of the nominal trajectory toward the nominal states. For this purpose, we use tubes to represent the neighborhood of the nominal trajectory as follow:

\textbf{Tube:} We define tube {\small$\mathcal{T}(k)$} as a neighborhood around the nominal trajectory {\small$\mathbf{x}^*$}, i.e., {\small$\mathbf{x}^*(k) \in \mathcal{T}(k), \ k=1,...,T$}. We will use basic simple sets to represent the tube at each time $k$. 
Examples are hyper-ellipsoid based tube defined as $\mathcal{T}(k)=\{ \mathbf{x} \in \mathbb{R}^{n_x}: \ (\mathbf{x}-\mathbf{x}^*(k))^TQ(k)(\mathbf{x}-\mathbf{x}^*(k)) \leq 1\}$, where $Q(k) \in \mathbb{R}^{n_x \times n_x}$ is the given positive definite matrix. A hyper-cube based tube defined as $\mathcal{T}(k)=\{ \mathbf{x} \in \mathbb{R}^{n_x}: \ |x_i-x_i^*(k)| \leq \epsilon(k), i=1,...n_x \}$, where $\epsilon(k) \in \mathbb{R}$.



\textbf{Time-Varying Feedback Controller:} Let {$\mathbf{\bar{x}}(k)=\mathbf{x}(k)-\mathbf{x}^*(k)$} be the error state vector. We look for the control input of the form {$\mathbf{u}(k)=\mathbf{\bar{u}}(k)+\mathbf{u}^*(k)$} where {$\mathbf{\bar{u}}(k)$} is the state feedback control in {$\mathbf{\bar{x}}(k)$}. 
More precisely, {$\mathbf{u}(k)=[u_{1}(k),...,u_{n_u}(k)]^T$} take the following form
\begin{align}\label{Con1}
& \mathbf{u}(k)=\mathbf{\bar{u}}(k)+\mathbf{u}^*(k) \\
& \bar{\bu}(k)= \mathbf{G}(k) \mathbf{\bar{x}}(k) \    \subeqn
\end{align}
\noindent where {$\mathbf{G}(k)=[{g_{ij}}(k)]_{(n_u, n_x)}$} is the feedback gain matrix of size $n_u \times n_x$ at time step $k$. 
The nonlinear state feedback case can be written out similarly and is considered in our Experiment A, but in this paper we mainly consider linear state feedback. 
The feedback control in \eqref{Con1} should satisfy the gain constraints as follows:
\begin{align}\label{Con2}
& {g}_{ij}(k)  \in [{a_i}_{j},{b_i}_j], \ i=1,...,n_u, \ j=1,...,n_x
\end{align}
\noindent where {${a_i}_j, {b_i}_j \in \mathbb{R}$} for $i=1,...,n_u$, $j=1,...,n_x$.
Now, we define the controller design problem for the stochastic nonlinear system of \eqref{sys} as follows.\\

\textbf{Controller Design Problem:} Given the probabilistic nonlinear system in \eqref{sys}, nominal trajectory {\small$\{\mathbf{x}^*, \mathbf{u}^*\}$}, and the tube $\mathcal{T}(k)$, we aim to design a state feedback of the form \eqref{Con1} satisfying the constraints \eqref{Con2} to follow the nominal trajectory and minimize the deviations due to  probabilistic uncertainties. The designed controller $\mathbf{u}(k)$ steers the deviated states $\mathbf{x}(k) \in \mathcal{T}(k)$ toward the given nominal trajectory. To obtain such controllers $\mathbf{u}(k),k=0,...,T$, we define the following optimization for each time step $k$:

\begin{align}
		&\min_{\mathbf{G}(k)}  && J(\mathbf{x}(k+1),\mathbf{x}^*(k+1))
		\label{P1}\\
		&\hbox{s.t.}\ \   &&  \mathbf{x}(k+1)=\mathbf{f}(\mathbf{x}(k),\mathbf{u}(k),\omega(k)) \tag{\ref{P1}a}\label{P1_1} \\
		&  &&\forall \mathbf{x}(k) \in \mathcal{T}(k) \tag{\ref{P1}b}\label{P1_2}\\
		& &&\omega(k) \sim p_{\omega_k}(\omega) \tag{\ref{P1}c}\label{P1_3} \\
		& &&\mathbf{u}(k)=\mathbf{\bar{u}}(k)+\mathbf{u}^*(k) \tag{\ref{P1}d}\\
		& &&\bar{\bu}(k)= \mathbf{G}(k) \mathbf{\bar{x}}(k) \tag{\ref{P1}e} \\
		& &&\mathbf{\bar{x}}(k)=\mathbf{x}(k)-\mathbf{x}^*(k) \tag{\ref{P1}f}\\
		& &&{\mathbf{G}(k)=[{g_{ij}}(k)]_{(n_u, n_x)}}  \tag{\ref{P1}g}\\
		& && {g_i}_j(k)  \in [{a_{i}}_j,{b_{i}}_j], \ i=1,...,n_u, j=1,...,n_x \tag{\ref{P1}h}
\end{align}
where $J$ is the objective function that measures the deviation of states $\mathbf{x}(k+1)$ from the given nominal state $\mathbf{x}^*(k+1)$ for all $\mathbf{x}(k) \in \mathcal{T}(k)$ in the presence of uncertainty $\omega(k)$. 
Problem in \eqref{P1} is a probabilistic nonlinear optimization that requires uncertainty propagation through the probabilistic nonlinear system of the robot to characterize the deviation of states from the given nominal trajectory. Hence, it is computationally challenging and hard. To solve the optimization in \eqref{P1}, we perform the following steps:

\textbf{i)} In the uncertainty minimization step, we define a proper objective function $J$ to measure the deviation of the states of the stochastic system form the given trajectory. We define $J$ in terms of the moments of the probability distribution of the states.

\textbf{ii)} In the nonlinear uncertainty propagation step, we obtain moments of the probability distributions of the states $\mathbf{x}(k+1)$ in terms of the control input $\mathbf{u}(k)$ in the presence of uncertainties $\omega(k)$ and $\mathbf{x}(k) \in \mathcal{T}(k)$. In this way, we are able to control the moments of the states using the control input $\mathbf{u}(k)$ to minimize the objective function $J$.

\textbf{iii)} In the final step, we use the moments obtained in steps (i) and (ii) to transform the probabilistic optimization \eqref{P1} into a deterministic optimization in terms of the moments and control input. Hence, we are able to use the off-the-shelf optimization solvers to obtain the control gains.

\section{Uncertainty Minimization}\label{sec_deviation}

\begin{figure}
    \centering
    \includegraphics[width=0.23\textwidth]{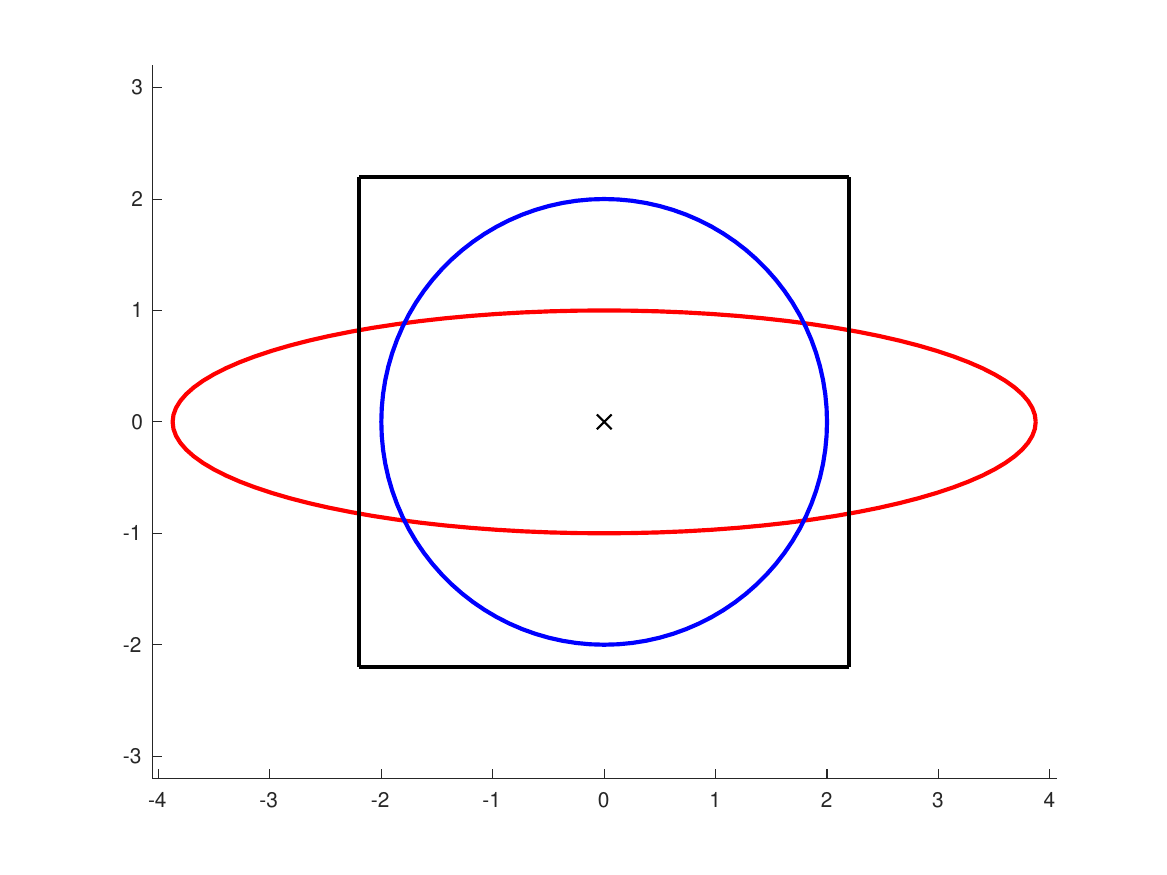}
    \includegraphics[width=0.23\textwidth]{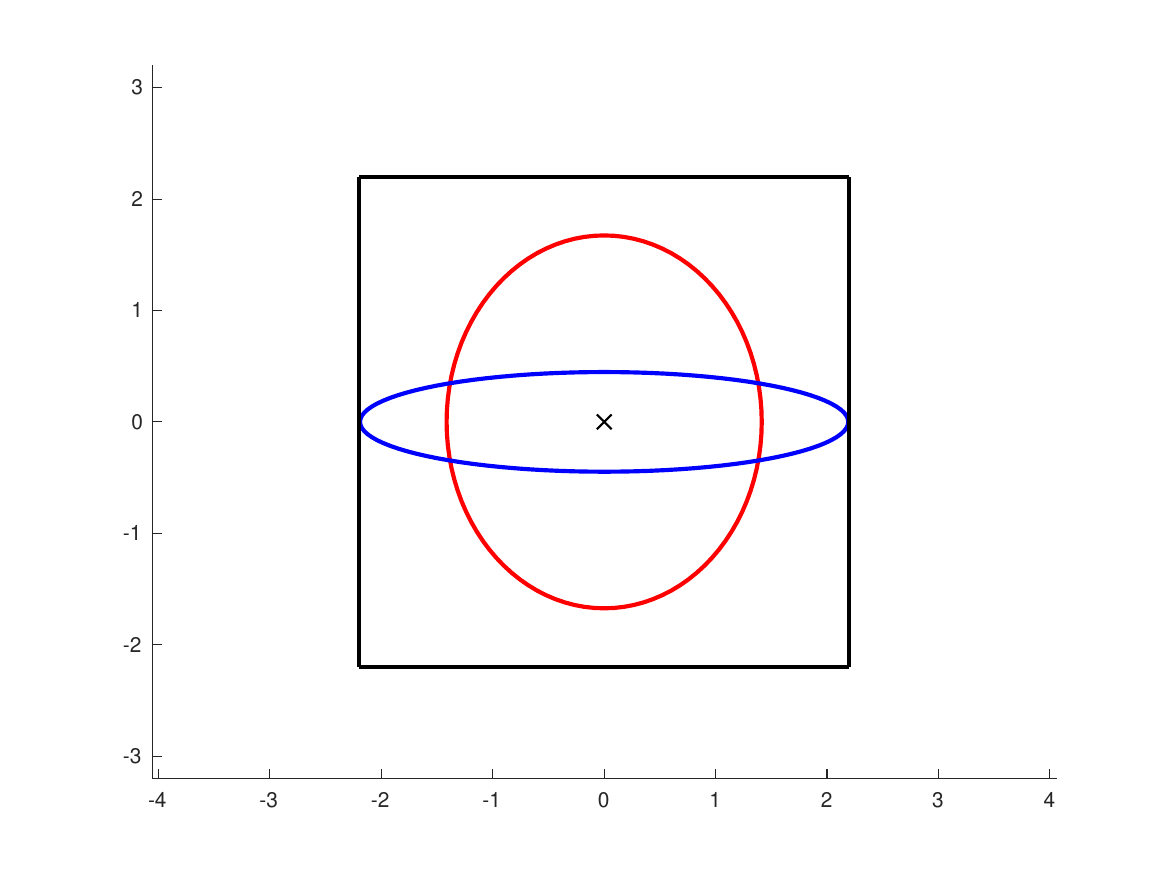}
    \caption{Error ellipses. Left: The red error ellipse corresponds to the covariance matrix with eigenvalues $\lambda_1 = 15$ and $\lambda_2 = 1$. The blue error ellipse corresponds to the covariance matrix with eigenvalues $\lambda_1 = 4$ and $\lambda_2 = 4$. The black box is the region where the user want the state to stay inside with high probability. Although the red ellipse corresponds to smaller determinant, the blue ellipse is more favourable to the user. Right: The red error ellipse corresponds to the covariance matrix with eigenvalues $\lambda_1 = 2.8$ and $\lambda_2 = 2$. The blue error ellipse corresponds to the covariance matrix with eigenvalues $\lambda_1 = 4.8$ and $\lambda_2 = 0.2$. Although the red ellipse corresponds to smaller trace, the blue ellipse can be more favourable to the user.}
    \label{fig:eigen}
\end{figure}

In this section, we define the objective function of optimization \eqref{P1} to capture the deviations of the probabilistic states of the system from the given nominal trajectory. Existing methods based on linearization around nominal trajectory and local Gaussian models usually use expected LQR cost as the objective:
\begin{align*}
    J = \mathbb{E}\left(\sum_{t = 0}^{N-1} \bx_t^\top Q \bx_t + \bu_t^\top R \bu_t + \bx_N^\top Q_f \bx_N \right)
\end{align*}

We take a different approach by directly looking at uncertainties of the states of the stochastic system. 
Given the state distribution at the current time step, we try to minimize the uncertainty of the states in the next time step. 
The eigenvalues of the covariance matrix $\mathbf{C}$ centered at the nominal trajectory capture the uncertainty or the deviation of states in the direction of eigenvectors. The larger the eigenvalue, the more uncertainty in the corresponding eigenvector direction. 
Geometrically, the square roots of the eigenvalues of the covariance matrix are proportional to the axis lengths of the error ellipse. 
Another natural objective would be to minimize the trace or the determinant of the covariance matrix $\mathbf{C}$. 
The trace is the sum of all eigenvalues and the determinant is the product of all eigenvalues. 
If the objective is to minimize determinant, then sometimes the resulting covariance matrix might represent an elongated error ellipse. 
For example in the 2-dimensional case, a covariance matrix $\mathbf{C}_1$ with eigenvalues $\lambda_1 = 15$ and $\lambda_2 = 1$ has smaller determinant than a covariance matrix $\mathbf{C}_2$ with eigenvalues $\lambda_1 = 4$ and $\lambda_2 = 4$, but $\mathbf{C}_1$ is less favourable than $\mathbf{C}_2$, because the deviation in one direction is too large (\Cref{fig:eigen} Left). 
Similarly, if the objective is the trace, then sometimes the optimization solver would prefer a covariance matrix $\mathbf{C}_1$ with eigenvalues $\lambda_1 = 2.8$ and $\lambda_2 = 2$ to a covariance matrix $\mathbf{C}_2$ with eigenvalues $\lambda_1 = 4.8$ and $\lambda_2 = 0.2$, although the later can be more favourable to users (\Cref{fig:eigen} Right). 

Under this consideration, we choose the weighted sum of the trace and the determinant as the objective
\begin{align*}
    J = w_1 \text{tr}(\mathbf{C}) + w_2 \det(\mathbf{C})
\end{align*}
where the weights $w_1$ and $w_2$ are hyperparameters that users can tune. 
When the resulting optimization problem is nonconvex, we also use multiple initializations to the optimization problem to get a larger variety of solutions, from which we are able to pick the favorite solution.






\section{Nonlinear Uncertainty Propagation}\label{sec_uncer_prop}
According to Section \ref{sec_deviation}, in order to minimize the deviations of the states of the stochastic system \eqref{sys}, we need the first and second order moments information of the states (covariance matrix). 
In this section, we describe the first and second order moments of states in terms of the control inputs. More precisely, we describe the first and second order moments of states at time $k+1$ in terms of the control inputs at time $k$, i.e., $\mathbb{E}[\mathbf{x}(k+1)]=\mathbb{E}[\mathbf{f}(\mathbf{x}(k),\mathbf{u}(k),\mathbf{\omega}(k))]$ and $\mathbb{E}[\mathbf{x}^2(k+1)]=\mathbb{E}[\mathbf{f}^2(\mathbf{x}(k),\mathbf{u}(k),\mathbf{\omega}(k))]$. 
This allows us to use control input $\mathbf{u}(k)$ to control the moments $\mathbb{E}[\mathbf{x}(k+1)]$ and $\mathbb{E}[\mathbf{x}^2(k+1)]$ and, therefore, minimize the deviations of the states. 

Given that function $\mathbf{f}$ in \eqref{sys} contains standard, trigonometric, and mixed trigonometric polynomial functions, to compute $\mathbb{E}[\mathbf{f}(\mathbf{x}(k),\mathbf{u}(k),\mathbf{\omega}(k))]$ and $\mathbb{E}[\mathbf{f}^2(\mathbf{x}(k),\mathbf{u}(k),\mathbf{\omega}(k))]$, we need to obtain the expected values of standard, trigonometric, and mixed trigonometric polynomial functions of uncertainties $\mathbf{x}(k)$ and $\omega(k)$.

We perform exact uncertainty propagation using characteristic function of random variables. The characteristic function is the Fourier transform of the probability density function of a random variable. For any random variable $x$, the characteristic function  always exists \cite{Char_1} and is defined as:
\begin{align} \label{CF}
    \Phi_{x}(t) = \mathbb{E}[e^{itx}]
\end{align}

Consider general mixed trigonometric polynomial moments of the form $\Exp[h]$, where $h$ is a product of $x^\alpha$, $\cos^{\beta_1}(c_1x)$, $\ldots$, $\cos^{\beta_p}(c_px)$, $\sin^{\gamma_1}(s_1x)$, $\ldots$, $\sin^{\gamma_q}(s_qx)$. This is more general than any case considered in \cite{jasour2021moment}. 
Instead of developing closed-form formulas, we develop the following algorithm to solve the problem. \\
\textbf{Algorithm Sketch:} Notice that 
\begin{align*}
    \cos^{\beta_j}(c_jx) &= \left({1\over 2}(e^{ic_jx} + e^{-ic_jx})\right)^{\beta_j}\\
    \sin^{\gamma_j}(s_jx) &= \left({1\over 2i}(e^{is_jx} - e^{-is_jx})\right)^{\gamma_j}
\end{align*}
So each $\cos^{\beta_j}(c_jx)$ and $\sin^{\gamma_j}(s_jx)$ can be written as a sum of terms of the form $ae^{ibx}$, where $a$ and $b$ are constants. 
The product of all sines and cosines is therefore also a sum of terms of the form $ae^{ibx}$. So the product $h$ is a sum of terms of the form $ax^\alpha e^{ibx}$.
By linearity of the expectation, $\Exp[h]$ is a sum of terms of the form $a\Exp[x^\alpha e^{ibx}]$, which can be computed as ${\partial^\alpha \over \partial t^\alpha} \Phi_x(t)|_{t=b}$ up to a constant. \qed

Using the defined moments, we address the nonlinear uncertainty propagation problem as follows. Given the stochastic nonlinear system in \eqref{sys}, probability distribution of uncertainty $\omega(k)$, and tube $\mathcal{T}(k)$, we obtain the moments of $\mathbf{x}(k+1)$ in the presence of uncertainties $\omega(k) \sim p_{\omega_k}(\omega)$ and $\mathbf{x}(k) \in \mathcal{T}(k)$.
To capture the deviations of $\mathbf{x}(k+1)$ for all the states $\mathbf{x}(k) \in \mathcal{T}(k)$, we use a uniform distribution over $\mathcal{T}(k)$ to represent $\mathbf{x}(k)$, i.e.,  $\mathbf{x}(k) \sim Uniform(\mathcal{T}(k))$. Then, the following holds true:
\begin{equation}\label{mom_prop}
   \mathbb{E}[\mathbf{x}^{\alpha}(k+1)]= \mathbf{F}_{\alpha}(\Phi_{\mathbf{x}(k)}(t),\Phi_{\mathbf{\omega}(k)}(t),\mathbf{G}(k))
\end{equation}
where $\mathbb{E}[\mathbf{x}^{\alpha}(k+1)]$ is the moment of order $\alpha$ of uncertain state $\mathbf{x}(k+1)$, $\Phi_{\mathbf{\omega}(k)}(t)$ is the characteristic function of $\omega(k)$, $\Phi_{\mathbf{x}(k)}(t)$ is the characteristic function of the uniform distribution defined over $\mathcal{T}(k)$, and $\mathbf{G}(k)$ is the state-feedback gain matrix. $\mathbf{F}_{\alpha}$ is a deterministic function in terms of expected values of the nonlinear functions of uncertainties. Hence, $\mathbf{F}_{\alpha}$ is a deterministic function of
characteristic functions $\Phi_{\mathbf{x}(k)}(t)$ and $\Phi_{\mathbf{\omega}(k)}(t)$ of the uncertainties and control input gains. \\

\section{Deterministic Optimization}
 
 In this section, we transform the optimization problem in \eqref{P1} into a deterministic optimization in terms of the moments of the probability distributions of the states and state feedback gains as follows:
 
\begin{align}
		&\min_{\mathbf{G}(k)} &&  w_1 \text{tr}(\mathbf{C}) + w_2 \det(\mathbf{C})
		\label{P2}\\
	&\hbox{s.t.}\	&&  \mathbb{E}[\mathbf{x}^{\alpha}(k+1)]= \mathbf{F}_{\alpha}(\Phi_{\mathbf{x}(k)}(t),\Phi_{\mathbf{\omega}(k)}(t),\mathbf{G}(k)) \tag{\ref{P2}a}\label{P2_1}\\
		& && {\mathbf{G}(k)=[{g_{ij}}(k)]_{(n_u, n_x)}}  \tag{\ref{P2}b}\\
		&  && {g_i}_j(k)  \in [{a_{i}}_j,{b_{i}}_j], \ i=1,...,n_u, j=1,...,n_x \tag{\ref{P2}c}
\end{align}
where the objective function is in terms of first and second order moments (covariance matrix) of the state $\mathbf{x}(k+1)$ defined in Section \ref{sec_deviation}, and the constraint \eqref{P2_1} implies the nonlinear uncertainty propagation step in \eqref{P1_1}, \eqref{P1_2}, \eqref{P1_3} described in detail in Section \ref{sec_uncer_prop}. 
The obtained optimization problem is in general non-convex, except for some simple scenarios.
We can solve the optimization problem using off-the-shelf interior point solvers.\\

\section{Implementation and Numerical Results}
In this section, we provide numerical examples to show the performance of the provided method. The computations in this section were performed on a computer with Intel i7 2.6 GHz processors and 16 GB RAM.
We use Casadi package \cite{andersson2019casadi} for Matlab as the wrapper to solve nonlinear optimization problems. 

\subsection{One-Dimensional Nonlinear Dynamics}
We first consider a one-dimensional stochastic nonlinear system \cite{yi2020nonlinear} defined as:
\begin{align}
    x(k+1) = x(k) + \cos(x(k))\Delta T + u(k)\Delta T + \alpha x^2(k) w(k)
\end{align}
where $\Delta T = 0.1$, the noise term $\alpha x^2(k) w(k)$ is state-dependent, and $\alpha$ is a constant set to be $0.1$.
We assume the initial distribution $x(0)$ has uniform distribution $x(0) \sim Uniform(-0.5,0.5)$, and the noise $w(k)$ has i.i.d. Gaussian distribution $w(k) \sim \mathcal{N}(\mu = 0, \sigma^2 = 0.01)$ for any $k$.
Given a sequence of nominal states $x^*(k)$ and controls $u^*(k)$ over the time horizon $N = 10$, where the nominal states $x^*(k)$ are the expected states at time $k$ by applying the sequence of nominal controls $u^*(k)$ starting from the initial distribution, we want the system to track the nominal trajectory $x^*(k)$ over the entire horizon.

Our method looks for the linear state feedback controller $u(k) = g(k) (x(k)-x^*(k)) + u^*(k)$, where $g(k)$ is the feedback gain to be determined. 
By plugging the controller into the system dynamics, we get
\begin{align}\label{eq:exp1_dyn2}
    x(k+1) &= x(k) + \cos(x(k))\Delta T \nonumber\\ 
    &+ (g(k) (x(k)-x^*(k)) + u^*(k))\Delta T + \alpha x^2(k) w(k)
\end{align}
The first moment of $x(k+1)$ is
\begin{align*}
    \mathbb{E}[x(k&+1)] = (1+g(k)\Delta T)\mathbb{E}[x(k)] + \Delta T \mathbb{E}[\cos(x(k))]\\
    &+\alpha \mathbb{E}[x^2(k)] \mathbb{E}[w(k)] +(-g(k)x^*(k)+u^*(k))\Delta T
\end{align*}
Similarly, the second moment of $x(k+1)$ can be expressed by squaring both sides of \Cref{eq:exp1_dyn2} and applying expectations on both sides.
The moments on the right-hand side of the expressions of $\mathbb{E}[x(k+1)]$ and $\mathbb{E}[x^2(k+1)]$ are of the form $\mathbb{E}[x^i(k)\cos^j(x(k))]$, where $0 \leq i \leq 4$ and $0 \leq j \leq 2$, and $\mathbb{E}[w^s(k)], s\leq 2$.
We consider the uniform distribution over the tube $\mathcal{T}(k) = \{x: |x - x^*(k)|\leq 0.5\}$.
We can compute the moments $\mathbb{E}[x^i(k)\cos^j(x(k))]$, obtaining the moments $\mathbb{E}[x(k+1)]$ and $\mathbb{E}[x^2(k+1)]$, and hence the variance of $x(k+1)$, which is a function of $g(k)$.
Finally, we use optimization solvers to solve for $g(k)$ by minimizing the variance of $x(k+1)$.
We have also considered the nonlinear state feedback controller $u(k) =g_1(k) (x(k)-x^*(k)) + g_2(k) (x(k)-x^*(k))^2 + u^*(k)$ and it turns out that the optimal controller that minimizes the variance of $x(k+1)$ has zero gain on the quadratic term, $g_2(k) = 0$. 
\begin{figure}
    \centering
\includegraphics[width=0.5\textwidth]{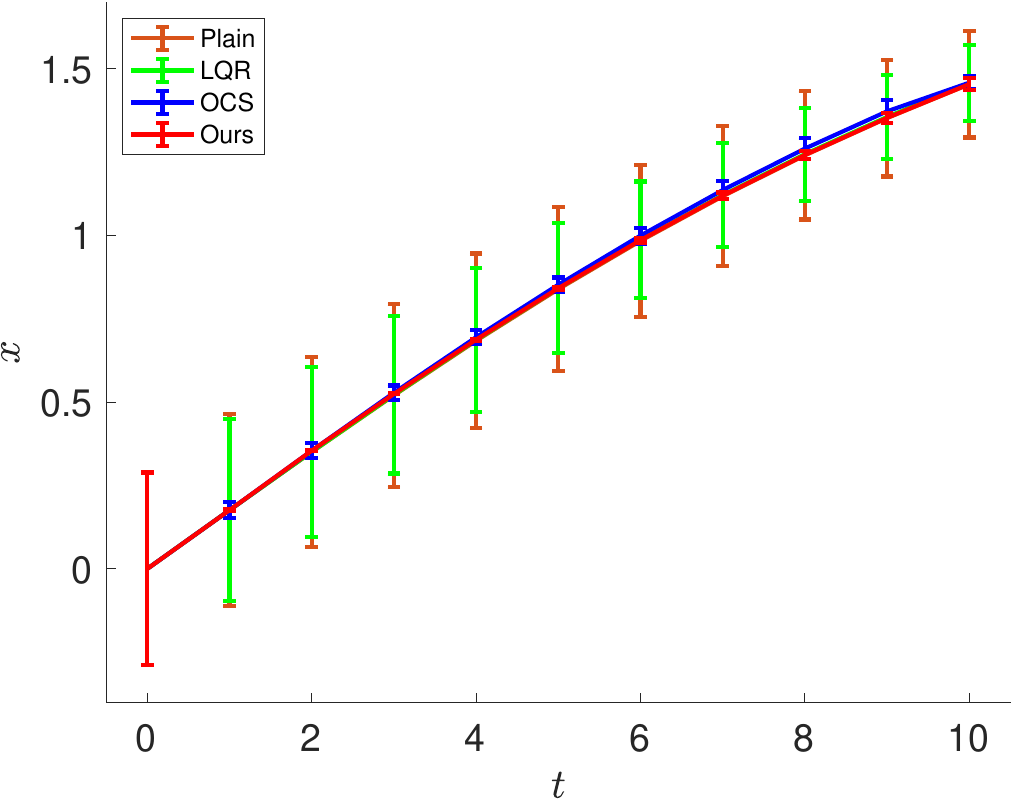}
    \caption{One-dimensional nonlinear dynamics: Mean and standard deviation of the state distribution over 10 time steps.}
    \label{fig:exp1}
\end{figure}
\begin{figure}
    \centering
\includegraphics[width=0.5\textwidth]{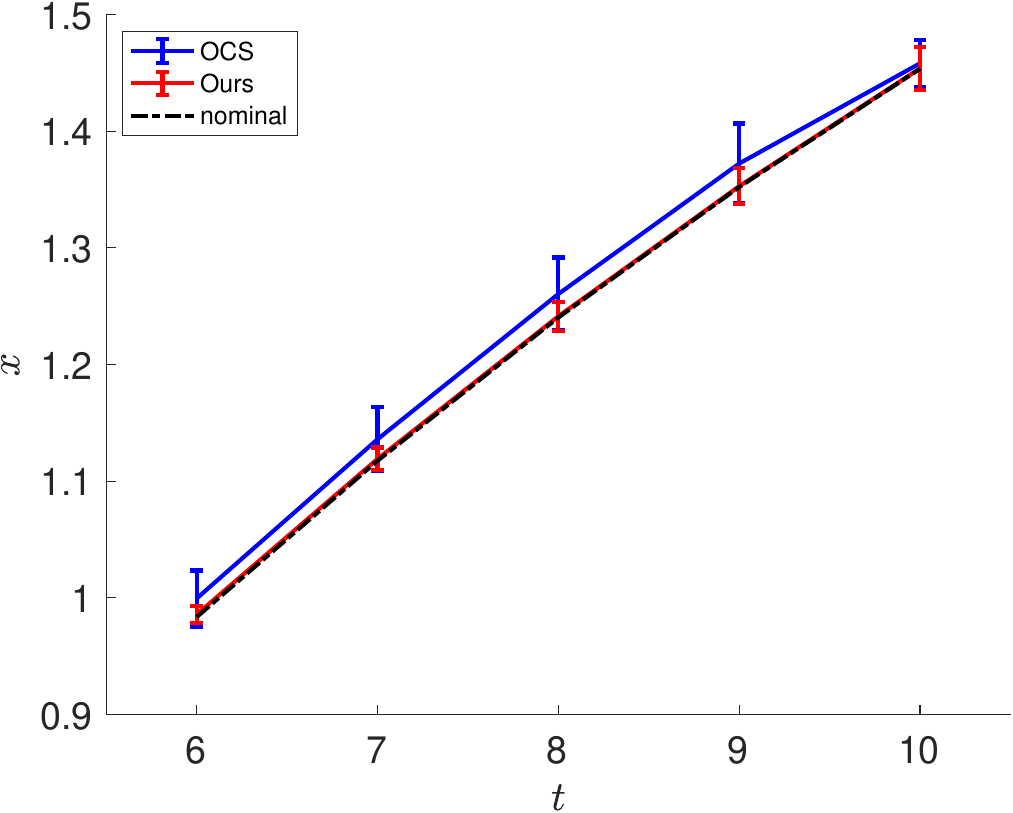}
    \caption{One-dimensional nonlinear dynamics: Zoom-in of \Cref{fig:exp1} between time steps 6 and 10, as well as the nominal trajectory. The mean of OCS deviates from the nominal trajectory. At time steps 6, $\ldots$, 9, the standard deviation of OCS is twice as large as ours.}
    \label{fig:exp1_2}
\end{figure}

\begin{figure*}[ht]
\includegraphics[width=0.32\textwidth]{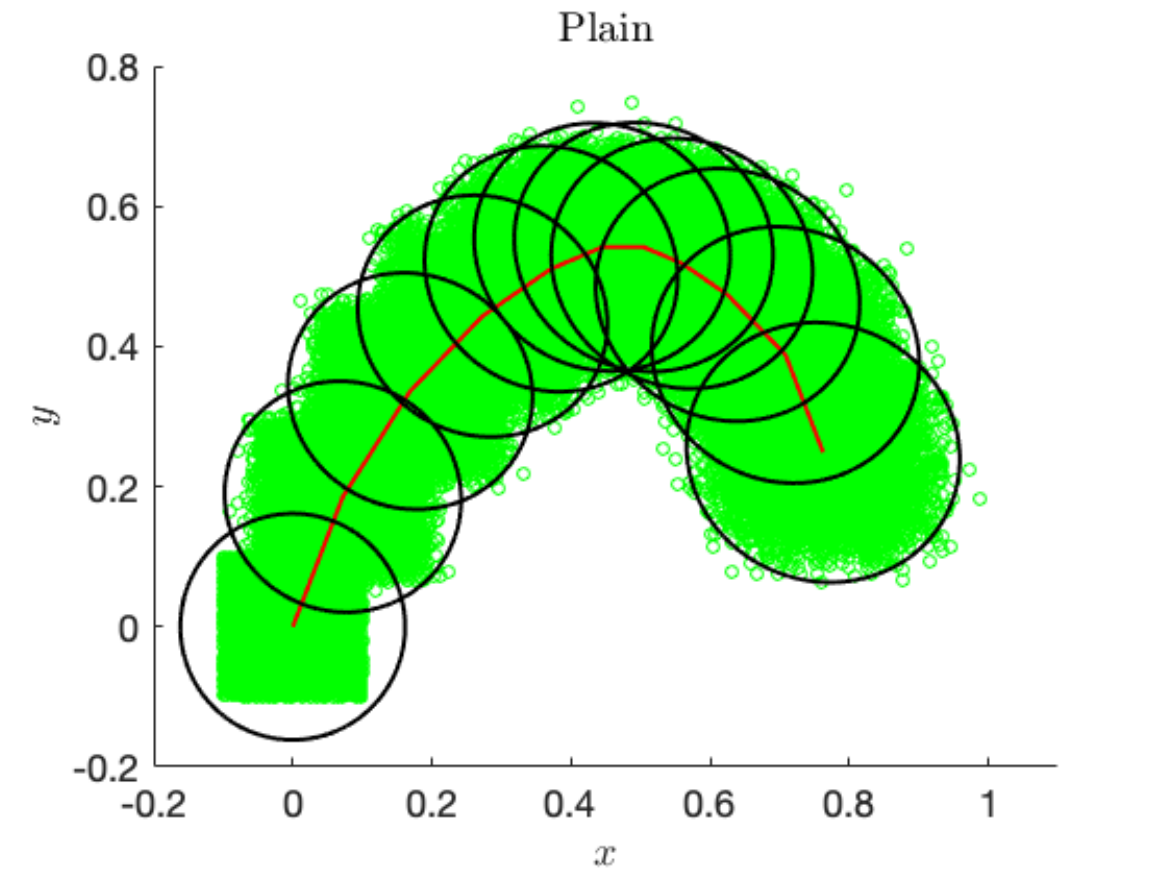}
\includegraphics[width=0.32\textwidth]{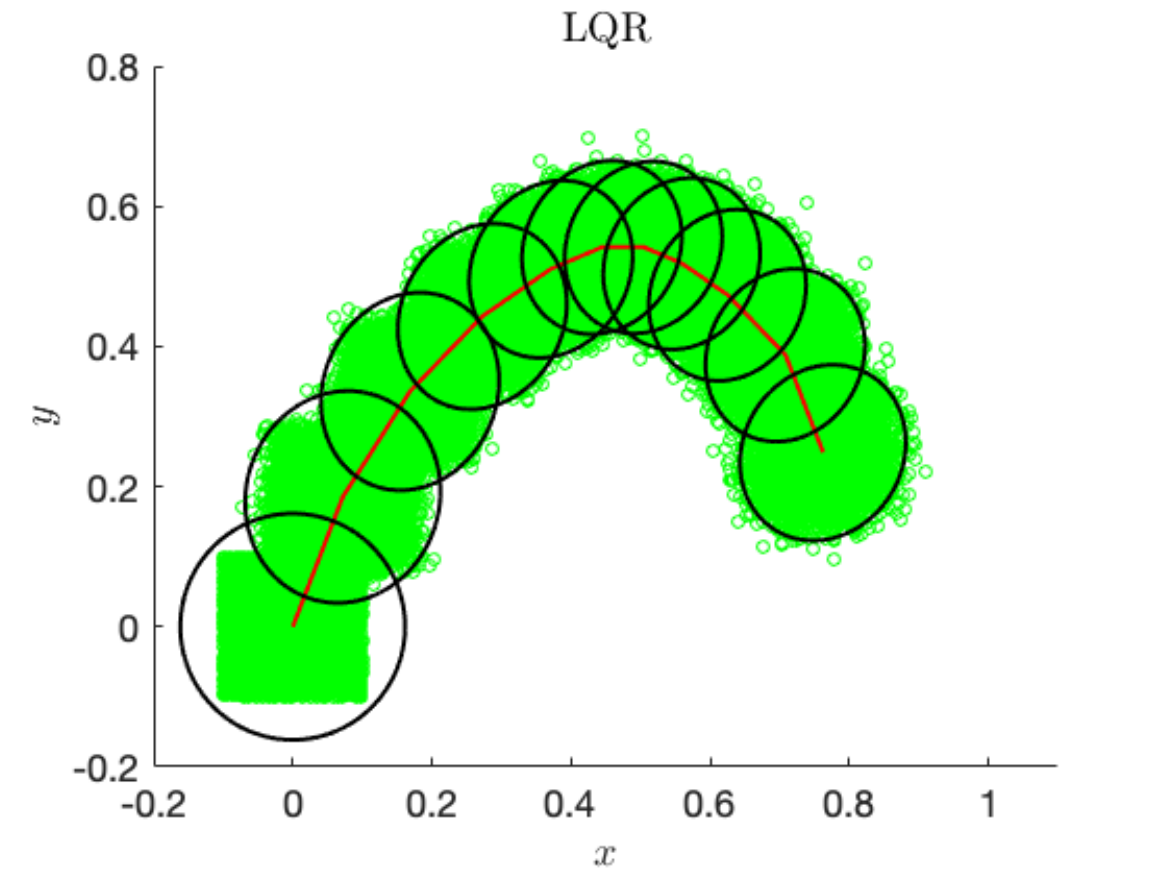}
\includegraphics[width=0.32\textwidth]{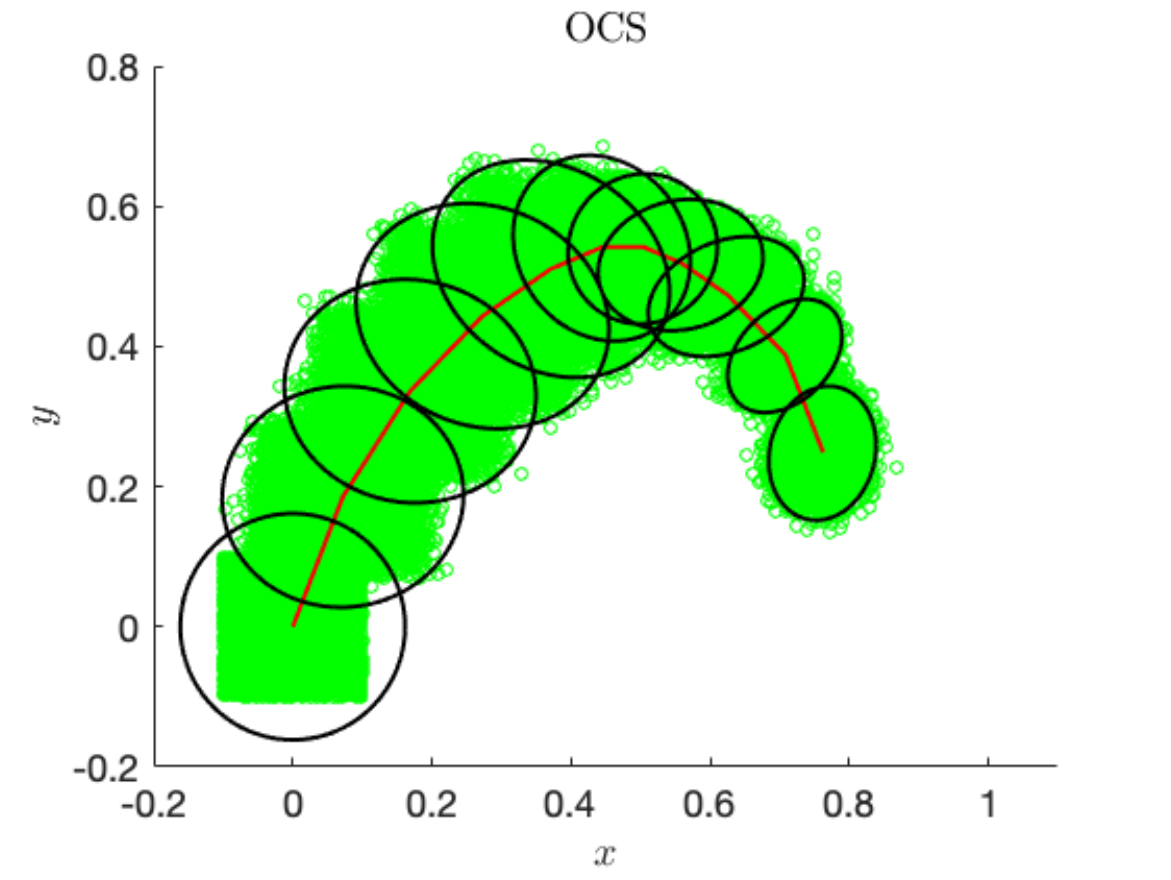}

\includegraphics[width=0.32\textwidth]{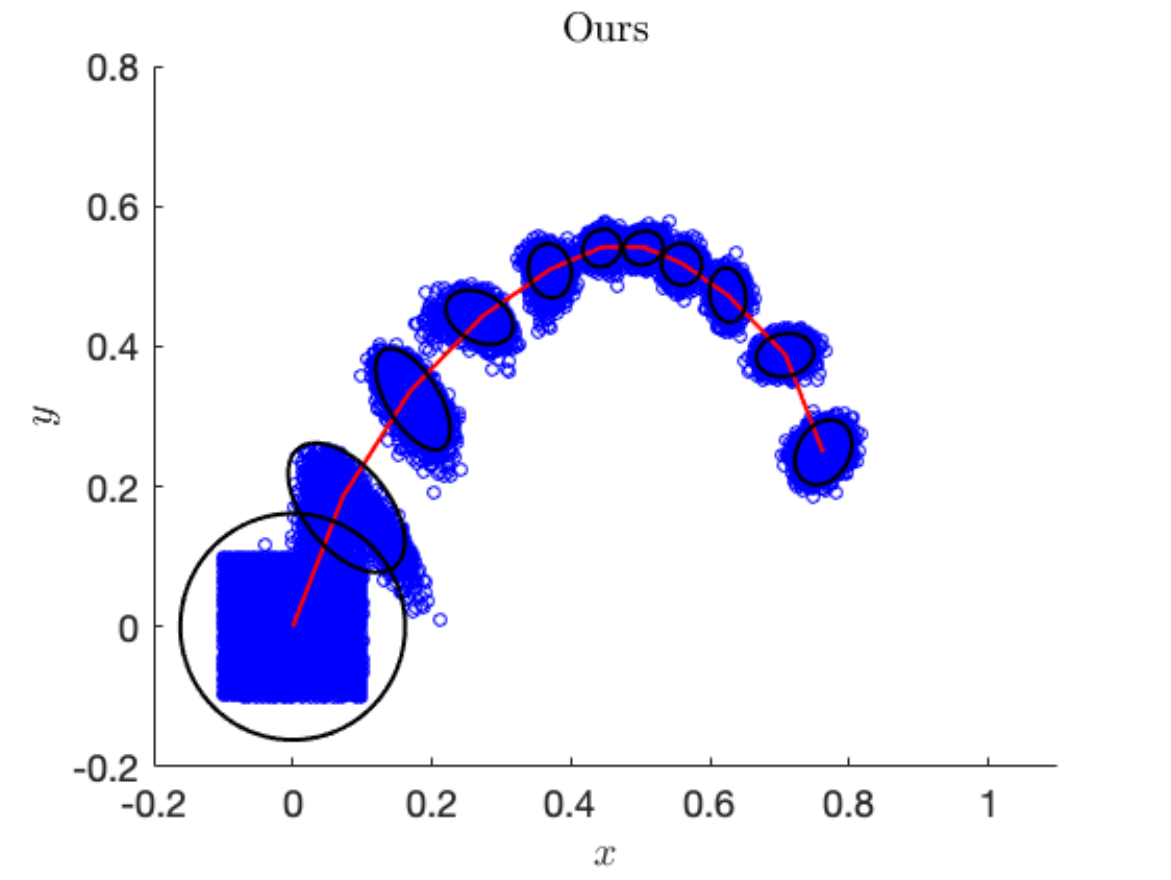}
\includegraphics[width=0.32\textwidth]{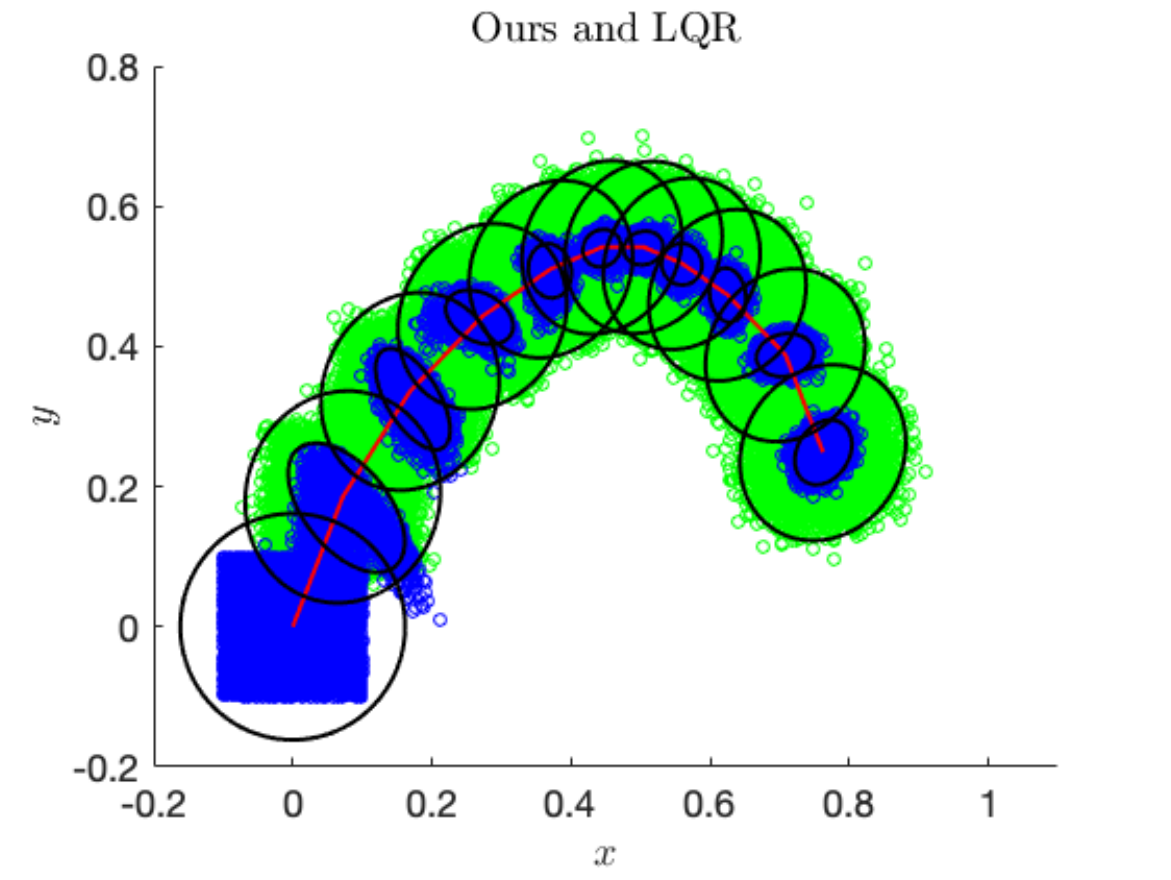}
\includegraphics[width=0.32\textwidth]{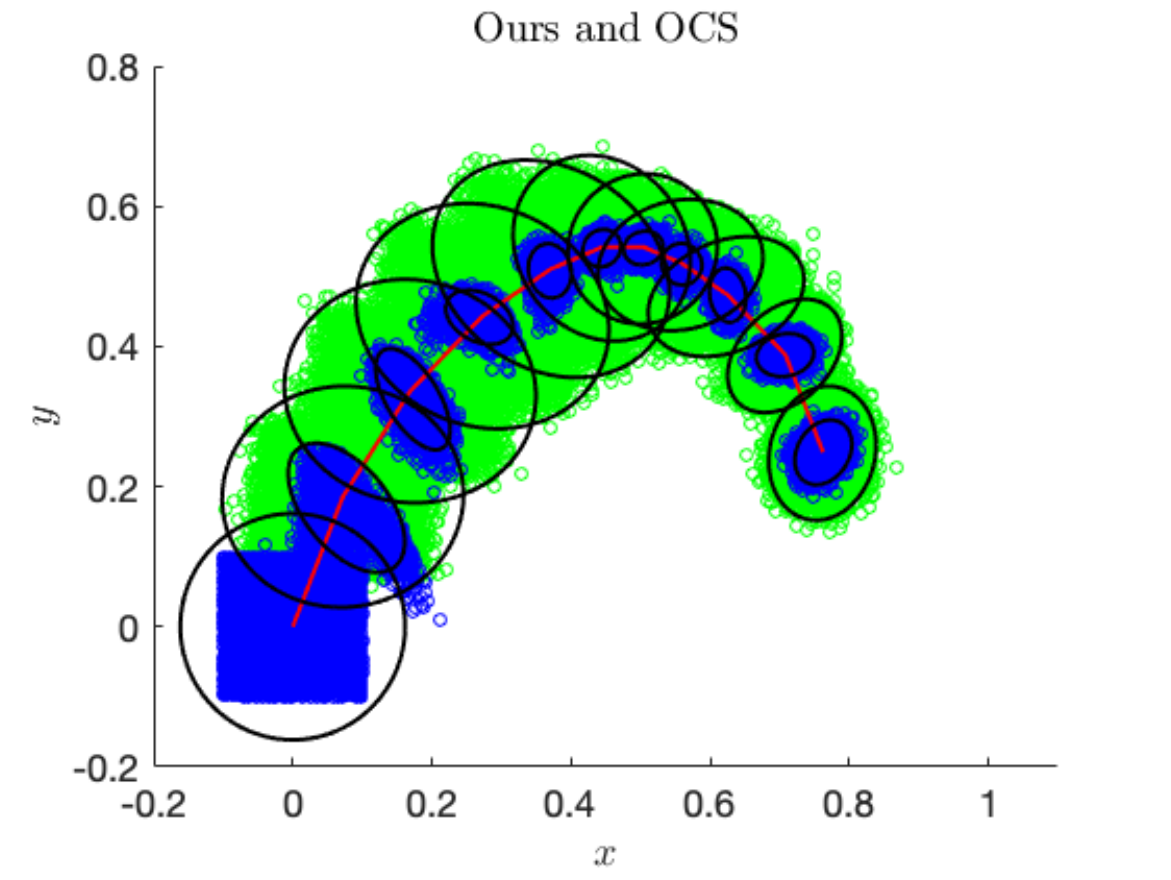}
\caption{Two-dimensional vehicle: State distributions over the entire horizon $N = 10$. Top, left to right: (1) Plain (without any controllers). (2) LQR controllers. (3) OCS controllers. Bottom, left to right: (1) Our controllers. (2) Our controllers (blue) laid on top of LQR controllers (green). (3) Our controllers (blue) laid on top of OCS controllers (green). The red line in all plots represents the nominal trajectory. The ellipses are $98\%$ confidence error ellipses.
LQR and OCS are tuned to give the best performance.
}
\label{fig:exp2}
\end{figure*}

\begin{table*}
\begin{center}
\begin{tabular}{ | c| c | c| c| c | c | c | c | c | c | c | } 
\hline 
max eigenvalues ($\times 10^{-3}$) & 1 & 2 & 3 & 4 & 5 & 6 & 7 & 8 & 9 & 10\\
\hline
Plain &
3.73 &
4.05 &
4.26 &
4.40 &
4.48 &
4.56 &
4.66 &
4.76 &
4.86 &
5.03\\
\hline
LQR & 
2.97 & 
2.66 & 
2.40 & 
2.21 & 
2.09 & 
2.05 & 
2.03 & 
2.02 & 
2.05 & 
2.16\\
\hline
OCS &
3.85 &
4.20 &
4.40 &
4.17 &
2.48 &
1.52 &
1.83 &
1.74 &
1.15 &
1.20 \\
\hline 
Ours & 
\textbf{1.52} & 
\textbf{0.85} & 
\textbf{0.34} & 
\textbf{0.19} & 
\textbf{0.10} & 
\textbf{0.10} & 
\textbf{0.11} & 
\textbf{0.19} & 
\textbf{0.22} & 
\textbf{0.31} \\
\hline
\end{tabular}

\vspace{1mm} 

\begin{tabular}{ | c| c | c| c| c | c | c | c | c | c | c | } 
\hline 
min eigenvalues ($\times 10^{-3}$) & 1 & 2 & 3 & 4 & 5 & 6 & 7 & 8 & 9 & 10\\
\hline
Plain &
3.44 &
3.54 &
3.65 &
3.76 &
3.85 &
3.90 &
3.93 &
4.00 &
4.14 &
4.30\\
\hline
LQR & 
2.47 & 
1.98 & 
1.73 & 
1.60 & 
1.55 & 
1.52 & 
1.51 & 
1.52 & 
1.57 & 
1.67 \\
\hline
OCS &
3.18 &
3.23 &
3.12 &
2.61 &
1.84 &
1.42 &
1.08 &
0.79 &
0.55 &
0.73 \\
\hline 
Ours & 
\textbf{0.46} &
\textbf{0.18} &
\textbf{0.15} &
\textbf{0.12} &
\textbf{0.08} &
\textbf{0.07} &
\textbf{0.11} &
\textbf{0.09} &
\textbf{0.12} &
\textbf{0.18} \\
\hline
\end{tabular}

\vspace{1mm} 

\begin{tabular}{ | c| c | c| c| c | c | c | c | c | c | c | } 
\hline 
trace ($\times 10^{-3}$) & 1 & 2 & 3 & 4 & 5 & 6 & 7 & 8 & 9 & 10\\
\hline
Plain &
7.17 &
7.59 &
7.91 &
8.16 &
8.32 &
8.46 &
8.59 &
8.76 &
9.00 &
9.33\\
\hline
LQR & 
5.44 & 
4.64 & 
4.13 & 
3.81 & 
3.65 & 
3.57 & 
3.54 & 
3.54 & 
3.62 & 
3.83\\
\hline
OCS &
7.03 & 
7.43 & 
7.52 & 
6.79 & 
4.32 & 
2.94 & 
2.92 & 
2.54 & 
1.70 & 
1.93 \\
\hline 
Ours & 
\textbf{1.99} & 
\textbf{1.03} & 
\textbf{0.49} & 
\textbf{0.32} & 
\textbf{0.19} & 
\textbf{0.18} & 
\textbf{0.22} & 
\textbf{0.28} & 
\textbf{0.34} & 
\textbf{0.49} \\
\hline
\end{tabular}

\vspace{1mm} 

\begin{tabular}{ | c| c | c| c| c | c | c | c | c | c | c | } 
\hline 
determinant ($\times 10^{-6}$) & 1 & 2 & 3 & 4 & 5 & 6 & 7 & 8 & 9 & 10\\
\hline
Plain &
12.82 &
14.33 &
15.56 &
16.53 &
17.22 &
17.78 &
18.33 &
19.03 &
20.12 &
21.61\\
\hline
LQR & 
7.34 & 
5.27 & 
4.15 & 
3.54 & 
3.25 & 
3.11 & 
3.06 & 
3.08 & 
3.21 & 
3.60\\
\hline
OCS &
12.23 &
13.55 &
13.73 &
10.90 &
4.56 &
2.16 &
1.99 &
1.38 &
0.63 &
0.88\\
\hline 
Ours & 
\textbf{0.71} & 
\textbf{0.15} & 
\textbf{0.05} & 
\textbf{0.02} & 
\textbf{0.01} & 
\textbf{0.01} & 
\textbf{0.01} & 
\textbf{0.02} & 
\textbf{0.03} & 
\textbf{0.06} \\
\hline
\end{tabular}
\caption{Two-dimensional vehicle: max/min eigenvalues, trace and determinant of the convariance matrices}
\label{table:exp2}
\end{center}
\end{table*}

To evaluate our methods, we sample 10,000,000 particles in the initial distribution $x(0) \sim Uniform(-0.5,0.5)$, and simulate forward in time with the nonlinear dynamics by applying our feedback controller. 
We plot the means and standard deviations over the entire horizon (\Cref{fig:exp1}). 
We compare our method with two other feedback control methods, LQR and OCS, whose parameters are tuned to give the best performance.

\noindent\textbf{LQR}: We linearize the system around the nominal trajectory while omitting the noise term, and get local time-varying linear systems around the nominal trajectory. Then we apply finite horizon linear-quadratic regulator (LQR) to the time-varying linear systems to get linear feedback controllers.

\noindent\textbf{OCS}: We linearize the system around the nominal trajectory (without omitting the noise term), and get local stochastic time-varying linear systems around the nominal trajectory. Then we apply optimal covariance steering (OCS) developed in \cite{okamoto2019optimal} to get linear feedback controllers.

\noindent\textbf{Plain}: We simulate the particle forward in dynamics using only the sequence of nominal controls. This shows us how large covariance can be without any feedback control. It also shows us the nominal trajectory.

From \Cref{fig:exp1} and \Cref{fig:exp1_2}, we observe that OCS and our method maintain small standard deviation over the entire trajectory, while LQR has large standard deviation. In general, our method's standard deviation is smaller than that of OCS. For example, as shown in \Cref{fig:exp1_2}, at time steps 6, $\ldots$, 9, the standard deviation of OCS is twice as large as ours. Our method's standard deviation increases slightly over the time horizon, because the noise term $\alpha x^2(k)w(k)$ depends on the state. When the state deviates more from 0, the noise is larger. The mean of our method and the mean of LQR coincide with the nominal state, while the mean of OCS deviates from the nominal state for more than half a standard deviation at time steps 6, $\ldots$, 9.

\begin{figure*}[ht]
\includegraphics[width=0.32\textwidth]{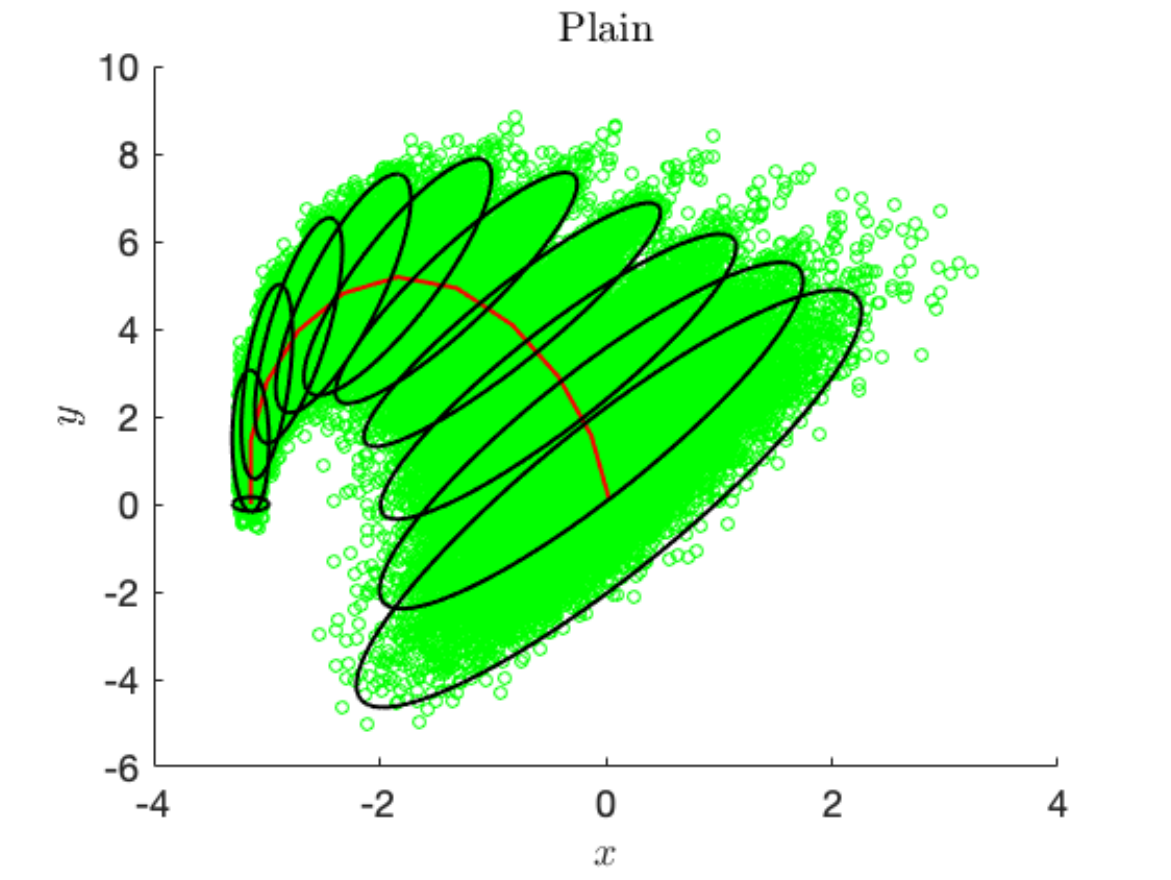}
\includegraphics[width=0.32\textwidth]{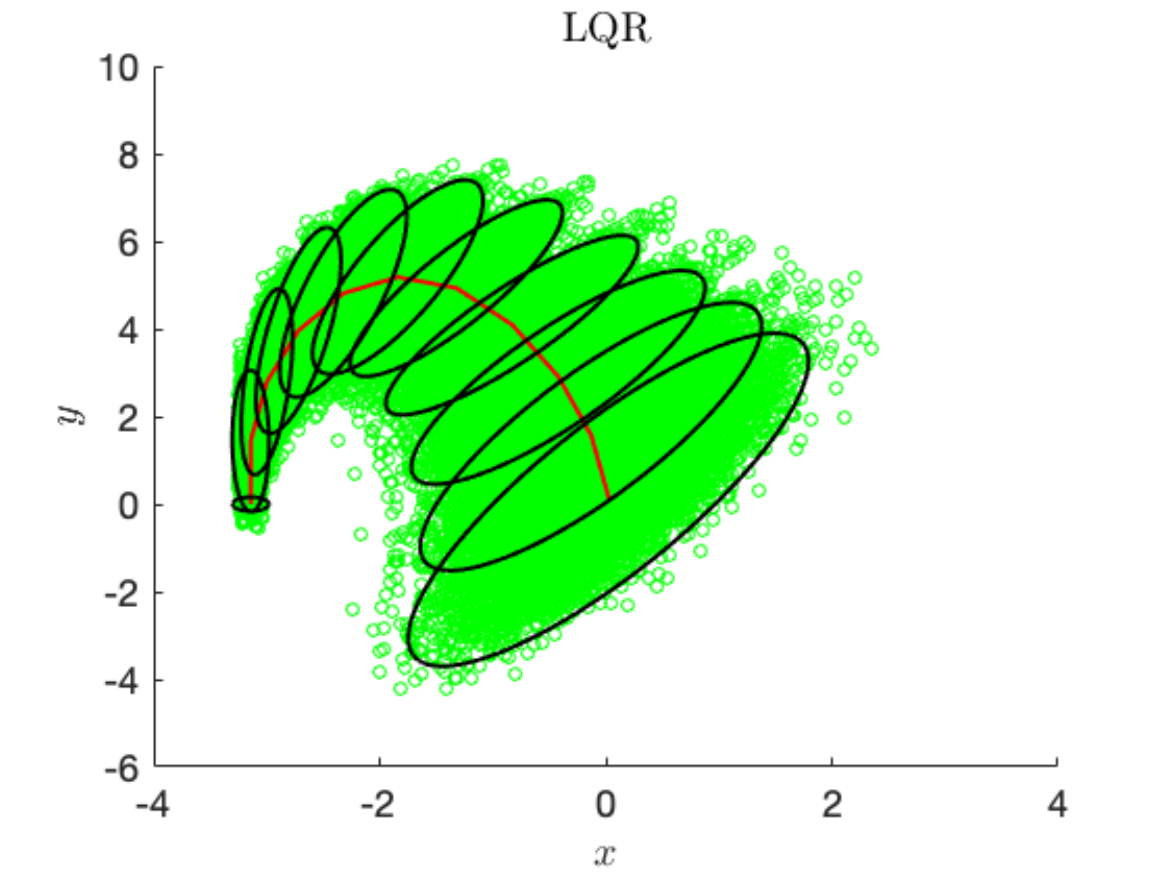}
\includegraphics[width=0.32\textwidth]{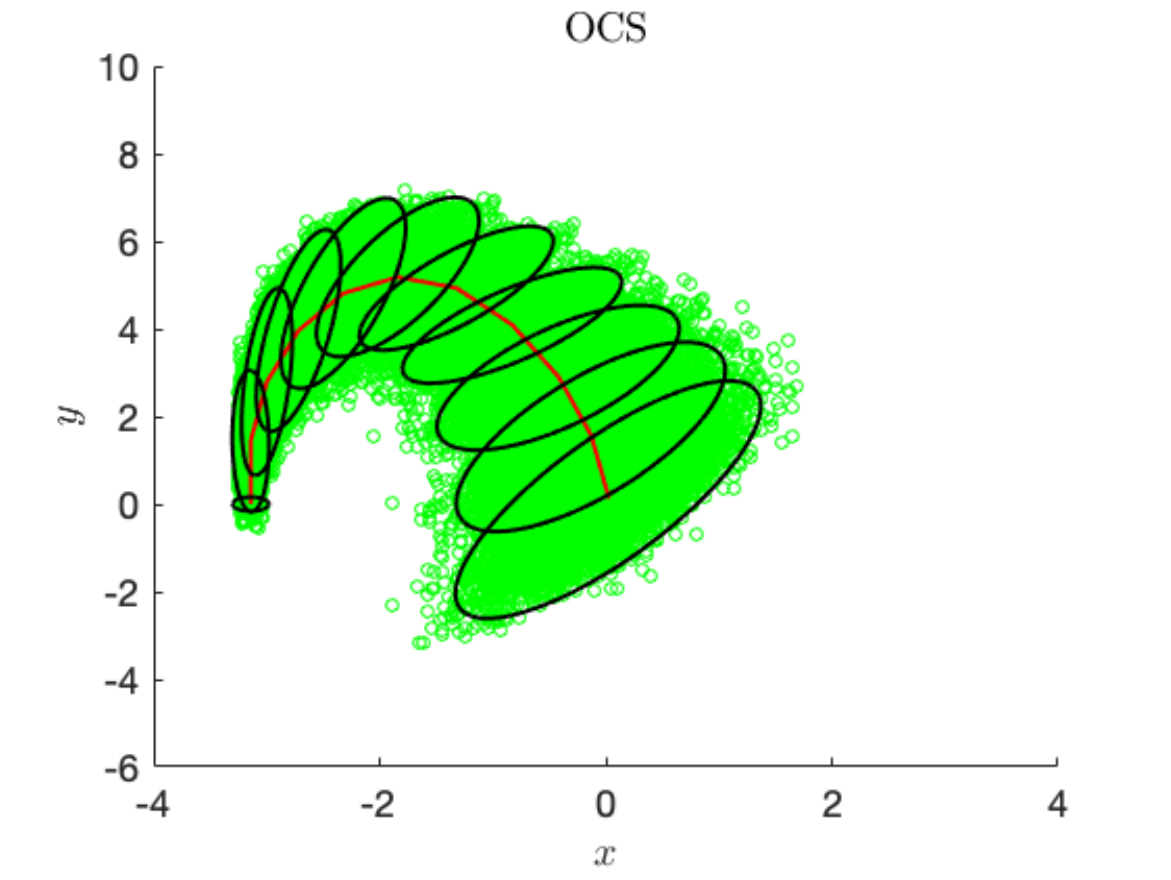}

\includegraphics[width=0.32\textwidth]{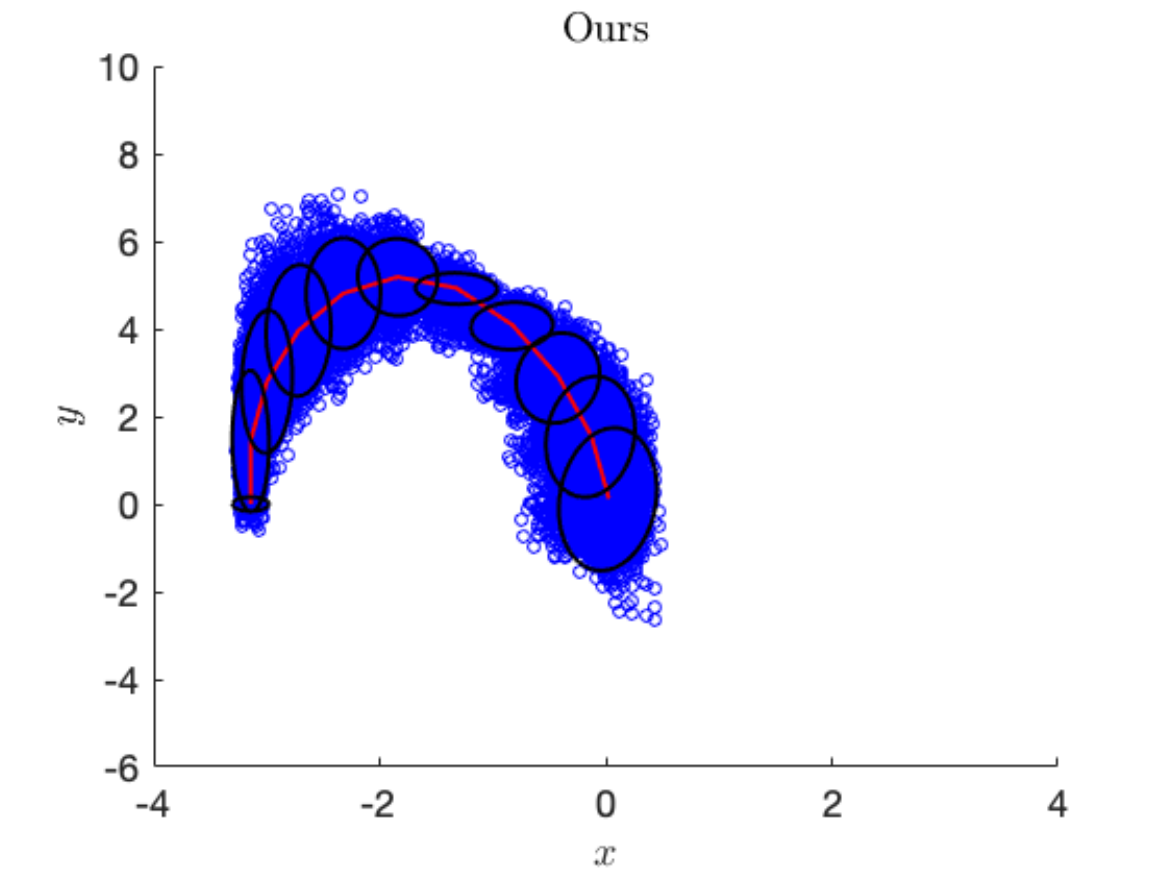}
\includegraphics[width=0.32\textwidth]{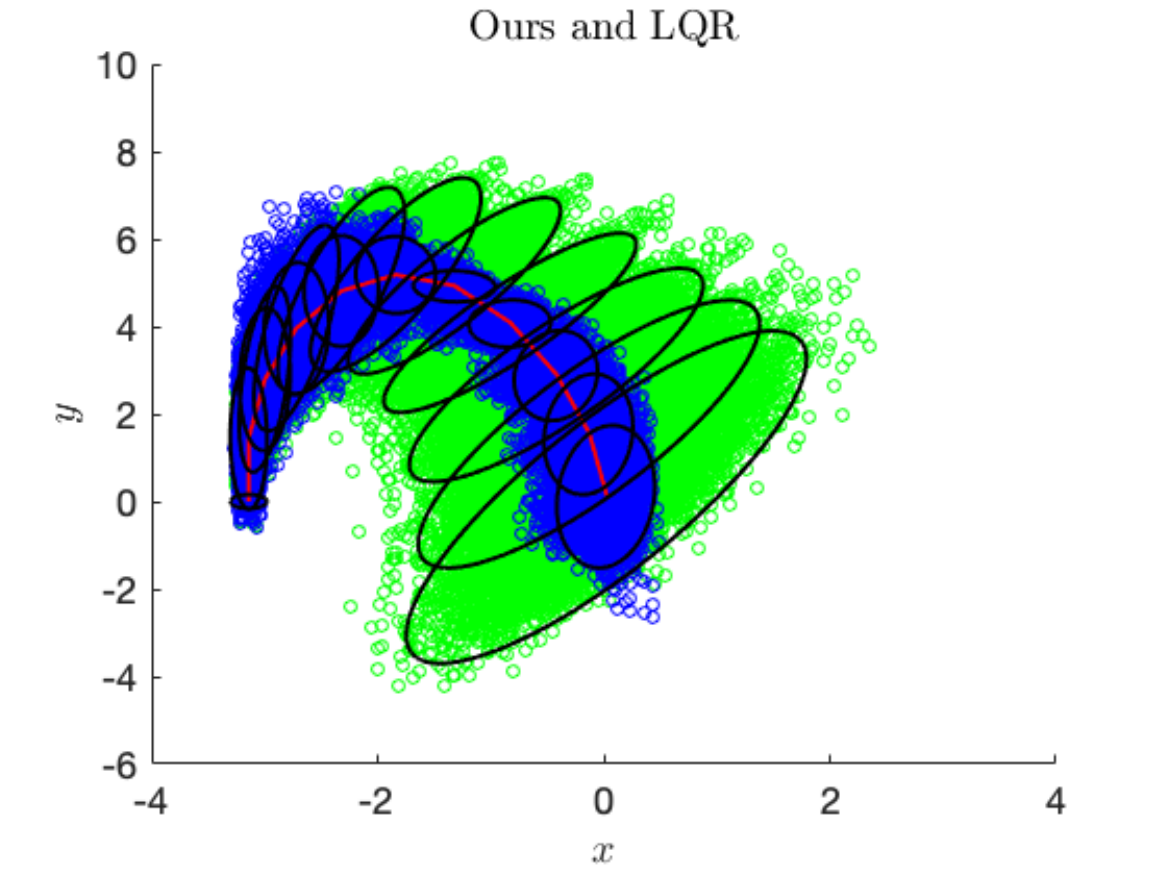}
\includegraphics[width=0.32\textwidth]{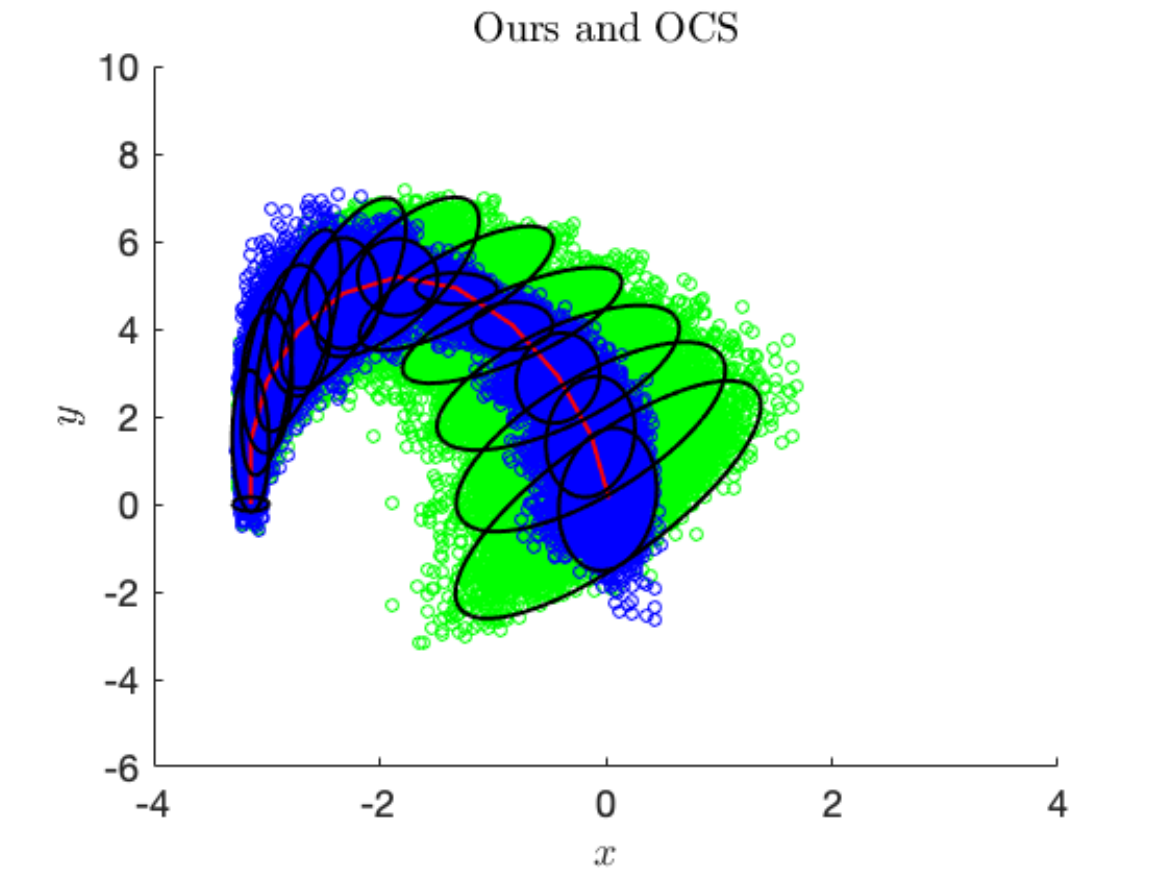}
\caption{Inverted pendulum: State distributions over the entire horizon $N = 10$. Top, left to right: (1) Plain (without any controllers). (2) LQR controllers. (3) OCS controllers. Bottom, left to right: (1) Our controllers. (2) Our controllers (blue) laid on top of LQR controllers (green). (3) Our controllers (blue) laid on top of OCS controllers (green). The red line in all plots represents the nominal trajectory. The ellipses are $98\%$ confidence error ellipses.
LQR and OCS are tuned to give the best performance.
}
\label{fig:exp3}
\end{figure*}

\subsection{Two-Dimensional Vehicle}

We consider the two-dimensional vehicle model in the form of 
\begin{align}\label{eq:exp2_dyn}
    x(k+1) &= x(k) + \Delta T(v(k) + w_v(k))\cos(\theta(k) + w_{\theta}(k))\nonumber\\
    y(k+1) &= y(k) + \Delta T(v(k) + w_v(k))\sin(\theta(k) + w_{\theta}(k))
\end{align}
The states $x(k)$ and $y(k)$ are the 2D coordinates of the vehicle. The control inputs  $v(k)$ and $\theta(k)$ are the velocity and the heading direction, respectively. $w_v(k)$ and $w_\theta(k)$ are noise with i.i.d. Gaussian distribution $w_v(k), w_\theta(k) \sim \mathcal{N}(\mu = 0, \sigma^2 =0.01)$. The time discretization is $\Delta T = 0.1$.
We have a nominal trajectory $(v^*(k), \theta^*(k)), (x^*(k), y^*(k))$ over the time horizon $N = 10$, with initial nominal state $(x^*(0), y^*(0))= (0,0)$, and goal nominal state $(x^*(10), y^*(10))= (0.7624,0.2488)$.
The initial state distribution is uniform $x(0), y(0) \sim Uniform(-0.1,0.1)$. 
We want to design state feedback controllers to track the nominal trajectory while minimizing the covariance of the states.

We search for linear state feedback controllers of the form
\begin{align}\label{eq:exp2_contr}
    \left[ \begin{matrix}v(k) \\ \theta(k) \end{matrix} \right]
    = \left[ \begin{matrix}g_{11}(k) & g_{12}(k) \\ g_{21}(k) & g_{22}(k) \end{matrix} \right]
    \left( \left[ \begin{matrix}x(k) \\ y(k) \end{matrix} \right]
    -\left[ \begin{matrix}x^*(k) \\ y^*(k) \end{matrix} \right]
    \right)+\left[ \begin{matrix}v^*(k) \\ \theta^*(k) \end{matrix} \right]
\end{align}
where $g_{11}(k), g_{12}(k), g_{21}(k), g_{22}(k)$ are feedback gains to be determined.
By plugging \Cref{eq:exp2_contr} into \Cref{eq:exp2_dyn}, we get expressions of $x(k+1)$ and $y(k+1)$ in terms of $x(k), y(k), w_v(k), w_\theta(k)$, as well as $g_{11}(k), g_{12}(k), g_{21}(k), g_{22}(k)$.
The moments of order 1 and 2, including $\mathbb{E}[x(k+1)], \mathbb{E}[y(k+1)], \mathbb{E}[x^2(k+1)], \mathbb{E}[x(k+1)y(k+1)]$, and $\mathbb{E}[y^2(k+1)]$, can be expressed as a sum of moments of the form $\mathbb{E}[h]$, where $h$ is a product of $x^p(k)$, $y^q(k)$, $\cos^{r_1}(g_{21}(k)x(k))$, $\sin^{r_2}(g_{21}(k)x(k))$,  $\cos^{s_1}(g_{22}(k)y(k))$, and $\sin^{s_2}(g_{22}(k)y(k))$. 
Since the noise $w_v(k)$ and $w_\theta(k)$ is independent of other random variables, their first two moments can be computed directly and be eliminated from the expressions without affecting other random variables.  
The number of all possible products $h$ can be bounded by a small number, because the powers in the product are bounded. For example, $p\leq 2$, $q\leq 2$, and there are at most two terms of sine or cosine involving $x(k)$ in the product $h$, and similar for $y(k)$. 
We consider the uniform distribution over the tube $\mathcal{T}(k) = \{(x,y): |x - x^*(k)|\leq 0.1, |y - y^*(k)|\leq 0.1\}$.
By computing all such moments $\mathbb{E}[h]$, the moments of $x(k+1)$ and $y(k+1)$ of order 1 and 2 can be expressed as nonlinear functions of feedback gains $g_{11}(k), g_{12}(k), g_{21}(k), g_{22}(k)$.
By minimizing a weighted sum of the determinant and the trace of the covariance matrix of $x(k+1)$ and $y(k+1)$, we get the values of the feedback gains $g_{11}(k), g_{12}(k), g_{21}(k), g_{22}(k)$.

To evaluate the obtained controllers, we sample 10,000,000 particles in the initial distribution $x(0), y(0) \sim Uniform(-0.1,0.1)$, and simulate forward under the nonlinear dynamics while applying our feedback controllers. We plot 10,000 particles in \Cref{fig:exp2} over the entire horizon $N = 10$. At each time step, we plot the $98\%$ confidence error ellipse \cite{errorellipse}, where the empirical mean and the empirical covariance matrix is computed from the 10,000,000 particles. We compare our method with LQR and OCS. \Cref{table:exp2} shows the numerical values, rounded to two decimal places, of the min/max eigenvalues, the trace, and the determinant of the covariance matrix at each time step.

From \Cref{fig:exp2} and \Cref{table:exp2}, we observe that our controllers achieve the smallest uncertainty of states over the entire trajectory among all comparison methods. Visually, our state distributions are the most concentrated around the nominal trajectory, and our error ellipses are the smallest in size, both in terms of volume and axis lengths, at each time step. Numerically, the min/max eigenvalues, the trace, and the determinant of the covariance matrix at each time step obtained by our method are much smaller than those of other methods. OCS controllers achieve relatively small uncertainty at the final time step, but have large uncertainty in earlier time steps. LQR steadily keeps the uncertainty at a certain level, which can be smaller than OCS at some time steps.  

\subsection{Inverted Pendulum}
We consider the inverted pendulum model \cite{yi2020nonlinear} in the form of 
\begin{align}
    x(k+1) &= x(k) + y(k) \Delta T\nonumber\\
    y(k+1) &= y(k) + 4\sin(x(k))\Delta T + u(k) \Delta T + \alpha u(k) w
\end{align}
where the state variables are $x(k) = \theta$ and $y(k) = \dot{\theta}$, the time discretization is $\Delta T = 0.1$, and the horizon is $N = 10$. The noise term is control dependent with the parameter $\alpha = 0.04$, and $w$ has standard normal distribution.
The initial state distribution is uniform over $[-0.1,0.1]^2$.
The nominal trajectory swings up the pendulum, going from $\theta = -180$ degrees to $\theta = 0$ degree, which corresponds to going from $(-\pi,0)$ to $(0,0)$ in state space. 
We design linear state feedback controllers of the form 
\begin{align*}
    u(k) = \left[\begin{matrix} g_1(k) & g_2(k) \end{matrix} \right] 
    \left( \left[ \begin{matrix}x(k) \\ y(k) \end{matrix} \right]
    -\left[ \begin{matrix}x^*(k) \\ y^*(k) \end{matrix} \right]
    \right)+\begin{matrix}u^*(k) \end{matrix}
\end{align*}
where $g_1(k)$ and $g_2(k)$ are feedback gains.
At each time step $k$, we consider the uniform distribution over the tube $\mathcal{T}(k) = \{(x,y): |x - x^*(k)|\leq 0.1, |y - y^*(k)|\leq 0.1\}$.
We solve optimization problems to minimize the state covariance of the next time step. 

To evaluate our controllers, we sample 10,000,000 particles in the initial distribution $Uniform([-0.1,0.1]^2)$, and simulate forward under the nonlinear dynamics while applying our feedback controllers. We plot 10,000 particles in \Cref{fig:exp3} over the entire horizon $N = 10$. At each time step, we plot the $98\%$ confidence error ellipse, where the empirical mean and the empirical covariance matrix is computed from the 10,000,000 particles. We compare our method with LQR and OCS.
Again we observe that our controllers achieve the smallest state uncertainty among all methods.

\section{Conclusion}
We  have  presented  a  method  to  design state feedback controllers for stochastic nonlinear robotic systems in the presence of arbitrary known probabilistic uncertainties. 
The obtained results show that the proposed method is promising and achieves the  smallest  uncertainty  of  states  among all comparison methods. 
A disadvantage of our approach is that it requires designing controllers per time step as in \cite{jasour2019sequential}, and this is the trade-off between performance and engineering effort. 
Nevertheless, our method could inspire more future work on control of stochastic nonlinear systems without linearization and Gaussian uncertainty assumptions. 

\bibliographystyle{IEEEtran}
\bibliography{references} 

\begin{thebibliography}{10}
\providecommand{\url}[1]{#1}
\csname url@samestyle\endcsname
\providecommand{\newblock}{\relax}
\providecommand{\bibinfo}[2]{#2}
\providecommand{\BIBentrySTDinterwordspacing}{\spaceskip=0pt\relax}
\providecommand{\BIBentryALTinterwordstretchfactor}{4}
\providecommand{\BIBentryALTinterwordspacing}{\spaceskip=\fontdimen2\font plus
\BIBentryALTinterwordstretchfactor\fontdimen3\font minus
  \fontdimen4\font\relax}
\providecommand{\BIBforeignlanguage}[2]{{%
\expandafter\ifx\csname l@#1\endcsname\relax
\typeout{** WARNING: IEEEtran.bst: No hyphenation pattern has been}%
\typeout{** loaded for the language `#1'. Using the pattern for}%
\typeout{** the default language instead.}%
\else
\language=\csname l@#1\endcsname
\fi
#2}}
\providecommand{\BIBdecl}{\relax}
\BIBdecl

\bibitem{aguiar2007trajectory}
A.~P. Aguiar and J.~P. Hespanha, ``Trajectory-tracking and path-following of
  underactuated autonomous vehicles with parametric modeling uncertainty,''
  \emph{IEEE transactions on automatic control}, vol.~52, no.~8, pp.
  1362--1379, 2007.

\bibitem{kuindersma2016optimization}
S.~Kuindersma, R.~Deits, M.~Fallon, A.~Valenzuela, H.~Dai, F.~Permenter,
  T.~Koolen, P.~Marion, and R.~Tedrake, ``Optimization-based locomotion
  planning, estimation, and control design for the atlas humanoid robot,''
  \emph{Autonomous robots}, vol.~40, no.~3, pp. 429--455, 2016.

\bibitem{han2020local}
W.~Han and R.~Tedrake, ``Local trajectory stabilization for dexterous
  manipulation via piecewise affine approximations,'' in \emph{2020 IEEE
  International Conference on Robotics and Automation (ICRA)}.\hskip 1em plus
  0.5em minus 0.4em\relax IEEE, 2020, pp. 8884--8891.

\bibitem{blackmore2009convex}
L.~Blackmore and M.~Ono, ``Convex chance constrained predictive control without
  sampling,'' in \emph{AIAA Guidance, Navigation, and Control Conference},
  2009, p. 5876.

\bibitem{blackmore2010probabilistic}
L.~Blackmore, M.~Ono, A.~Bektassov, and B.~C. Williams, ``A probabilistic
  particle-control approximation of chance-constrained stochastic predictive
  control,'' \emph{IEEE transactions on Robotics}, vol.~26, no.~3, pp.
  502--517, 2010.

\bibitem{okamoto2019optimal2}
K.~Okamoto and P.~Tsiotras, ``Optimal stochastic vehicle path planning using
  covariance steering,'' \emph{IEEE Robotics and Automation Letters}, vol.~4,
  no.~3, pp. 2276--2281, 2019.

\bibitem{okamoto2019optimal}
K.~Okamoto, ``Optimal covariance steering: Theory and its application to
  autonomous driving,'' Ph.D. dissertation, Georgia Institute of Technology,
  2019.

\bibitem{chen2015optimal}
Y.~Chen, T.~T. Georgiou, and M.~Pavon, ``Optimal steering of a linear
  stochastic system to a final probability distribution, part i,'' \emph{IEEE
  Transactions on Automatic Control}, vol.~61, no.~5, pp. 1158--1169, 2015.

\bibitem{chen2015optimal2}
------, ``Optimal steering of a linear stochastic system to a final probability
  distribution, part ii,'' \emph{IEEE Transactions on Automatic Control},
  vol.~61, no.~5, pp. 1170--1180, 2015.

\bibitem{chen2018optimal}
------, ``Optimal steering of a linear stochastic system to a final probability
  distribution—part iii,'' \emph{IEEE Transactions on Automatic Control},
  vol.~63, no.~9, pp. 3112--3118, 2018.

\bibitem{yi2020nonlinear}
Z.~Yi, Z.~Cao, E.~Theodorou, and Y.~Chen, ``Nonlinear covariance control via
  differential dynamic programming,'' in \emph{2020 American Control Conference
  (ACC)}.\hskip 1em plus 0.5em minus 0.4em\relax IEEE, 2020, pp. 3571--3576.

\bibitem{ridderhof2019nonlinear}
J.~Ridderhof, K.~Okamoto, and P.~Tsiotras, ``Nonlinear uncertainty control with
  iterative covariance steering,'' in \emph{2019 IEEE 58th Conference on
  Decision and Control (CDC)}.\hskip 1em plus 0.5em minus 0.4em\relax IEEE,
  2019, pp. 3484--3490.

\bibitem{tsolovikos2020nonlinear}
A.~Tsolovikos and E.~Bakolas, ``Nonlinear covariance steering using variational
  gaussian process predictive models,'' \emph{arXiv preprint arXiv:2010.00778},
  2020.

\bibitem{jasour2019sequential}
A.~M. Jasour and B.~C. Williams, ``Sequential chance optimization for flow-tube
  based control of probabilistic nonlinear systems,'' in \emph{2019 IEEE 58th
  Conference on Decision and Control (CDC)}.\hskip 1em plus 0.5em minus
  0.4em\relax IEEE, 2019, pp. 5392--5399.

\bibitem{han2022non}
W.~Han, A.~Jasour, and B.~Williams, ``Non-gaussian risk bounded trajectory
  optimization for stochastic nonlinear systems in uncertain environments,'' in
  \emph{2022 International Conference on Robotics and Automation (ICRA)}.\hskip
  1em plus 0.5em minus 0.4em\relax IEEE, 2022, pp. 11\,044--11\,050.

\bibitem{Tri_1}
G.~P{\'o}lya and G.~Szeg{\"o}, ``Polynomials and trigonometric polynomials,''
  in \emph{Problems and Theorems in Analysis}.\hskip 1em plus 0.5em minus
  0.4em\relax Springer, 1976, pp. 252--278.

\bibitem{jasour2021moment}
A.~Jasour, A.~Wang, and B.~C. Williams, ``Moment-based exact uncertainty
  propagation through nonlinear stochastic autonomous systems,'' \emph{arXiv
  preprint arXiv:2101.12490}, 2021.

\bibitem{Mixed_Tri_1}
T.~Lutovac, B.~Male{\v{s}}evi{\'c}, and C.~Mortici, ``The natural algorithmic
  approach of mixed trigonometric-polynomial problems,'' \emph{Journal of
  inequalities and applications}, vol. 2017, no.~1, pp. 1--16, 2017.

\bibitem{Mixed_Tri_2}
S.~Chen and Z.~Liu, ``Automated proof of mixed trigonometric-polynomial
  inequalities,'' \emph{Journal of Symbolic Computation}, vol. 101, pp.
  318--329, 2020.

\bibitem{Char_1}
E.~Cinlar and E.~{\dh}C{\i}nlar, \emph{Probability and stochastics}.\hskip 1em
  plus 0.5em minus 0.4em\relax Springer, 2011, vol. 261.

\bibitem{andersson2019casadi}
J.~A. Andersson, J.~Gillis, G.~Horn, J.~B. Rawlings, and M.~Diehl, ``Casadi: a
  software framework for nonlinear optimization and optimal control,''
  \emph{Mathematical Programming Computation}, vol.~11, no.~1, pp. 1--36, 2019.

\bibitem{errorellipse}
{AJ Johnson (2022)}, ``{error\_ellipse},''
  \url{https://www.mathworks.com/matlabcentral/fileexchange/4705-error_ellipse},
  2015, online; accessed 28 January 2022.

\end{thebibliography}


\begin{thebibliography}{1}




\bibitem{MOM1}
J. Jacod, and P. Protter, ”Probability Essentials”, Springer, Universitext book series, 2004.

\bibitem{SOS1}
P. A. Parrilo, ”Semidefinite programming relaxations for semialgebraic problems”, Mathematical Programming, vol. 96, pp. 293--320, 2003.


\bibitem{SOS2}
J. B. Lasserre, ”Global optimization with polynomials and the problem of moments”, SIAM Journal on Optimization, vol. 11, pp. 796--817, 2011.

\bibitem{SOS3}
M. Laurent, ”Sums of squares, moment matrices and optimization
over polynomials”, In: Putinar M., Sullivant S. (eds) Emerging
Applications of Algebraic Geometry. The IMA Volumes in Mathematics and its Applications, vol. 149, Springer, New York, 2009.


\bibitem{Yalmip_1}
Lofberg,"YALMIP : A Toolbox for Modeling and Optimization in MATLAB", yalmip.github.io/tutorial/sumofsquaresprogramming.



\bibitem{Glopti}
D. Henrion, J. B. Lasserre, J. Loefberg, "GloptiPoly 3: moments, optimization and semidefinite programming", Optimization Methods and Software, Vol. 24, Nos. 4--5, pp. 761--779, 2009.



\bibitem{SOS_time1}
A. A. Ahmadi and B. El Khadir, "Time-Varying Semidefinite Programs",to be appear in Mathematics of Operations Research, arXiv:1808.03994, 2019.

\bibitem{SOS_time2}
B.E Khadir, J.B Lasserre, V Sindhwani, "Piecewise-Linear Motion Planning amidst Static, Moving, or Morphing Obstacles", arXiv:2010.08167, 2020.

\bibitem{Contour}
A. Jasour, B. C. Williams, "Risk Contours Map for Risk Bounded Motion Planning under Perception Uncertainties", Robotics: Science and System (RSS), Germany, 2019.

\bibitem{Risk_Ind}
A. Jasour, A. Hofmann, and B. C. Williams, "Moment-Sum-Of-Squares Approach for Fast Risk Estimation in Uncertain Environments", IEEE Conference on Decision and Control (CDC), 2018.

\bibitem{Contour2}
 A. Jasour, ”Risk Aware and Robust Nonlinear Planning (rarnop)”, Course Notes for MIT 16.S498, rarnop.mit.edu, 2019.

\bibitem{Spot_1}
 M. M. Tobenkin, F. Permenter, A. Megretski, "spotless: Polynomial and Conic Optimization", github.com/spot-toolbox/spotless, 2013.

\end{thebibliography}

\end{document}